\documentclass[10pt,twocolumn,letterpaper]{article}
\newif\ifforceplain\forceplainfalse %

\newif\ifarxiv
\arxivtrue

\ifarxiv
\usepackage{cvpr}\cvprfinalcopy                         %
\else
\usepackage{cvpr}\cvprfinalcopy                        %
\fi

\makeatletter						%
\@namedef{ver@everyshi.sty}{}
\makeatother
\usepackage{tikz}

\usepackage{times}					%

\usepackage{epsfig}
\usepackage{graphicx}
\usepackage{amsmath}
\usepackage{amssymb}
\usepackage{array}
\usepackage{multirow}
\usepackage{fancyhdr}

\usepackage[mathscr]{eucal}
\usepackage{amsfonts}
\usepackage{color}
\usepackage{calc}
\usepackage[normalem]{ulem}  %

\ifarxiv
\newcommand{\ext}{jpg}  %
\else
\newcommand{\ext}{jpg}
\fi

\newif\iflualatex
\lualatexfalse
\iflualatex
	
	\usepackage[no-math]{fontspec}		%
\fi

\usepackage[pagebackref=true,breaklinks=true,letterpaper=true,colorlinks,bookmarks=false]{hyperref}

\newcommand{\arch}[1]{\textsc{#1}}

\def\clap#1{\hbox to 0pt{\hss #1\hss}}%
\definecolor{internationalkleinblue}{rgb}{0.0, 0.18, 0.65}	%
\definecolor{olive}{rgb}{0.5, 0.5, 0.0}
\definecolor{gray}{rgb}{0.7, 0.7, 0.7}
\definecolor{maroon}{rgb}{0.69, 0.19, 0.38}
\definecolor{celestialblue}{rgb}{0.29, 0.59, 0.82}
\definecolor{darkblue}{rgb}{0.19, 0.19, 0.62}
\definecolor{rust}{rgb}{0.72, 0.25, 0.05}
\definecolor{red}{rgb}{1.00, 0.00, 0.00}

\newcommand{\FINAL}[2][]{#2} %

\newcommand{\yy}{{\bf y}}
\newcommand{\zz}{{\bf z}}
\newcommand{\ww}{{\bf w}}
\newcommand{\tildeww}{{\bf\tilde w}}

\newcommand{\YY}{\mathcal{Y}}
\newcommand{\ZZ}{\mathcal{Z}}
\newcommand{\WW}{\mathcal{W}}
\newcommand{\MU}{\boldsymbol{\mu}}
\newcommand{\II}{\mathbf{I}}

\newcommand{\JJ}{{\bf J}}
\newcommand{\JJT}{{\bf J}^T}
\newcommand{\SSigma}{{\bf \Sigma}}
\newcommand{\real}{\mathbb{R}}
\newcommand{\expectation}{\mathbb{E}}
\newcommand{\tyy}{\tilde {\bf y}}

\newcommand{\h}{0mm}
\newcommand{\hh}{0mm}

\newcommand{\s}{\hphantom{0}}

\newcommand{\figdroplets}{
\begin{figure*}[t]
\renewcommand{\h}{0.163\linewidth}
\includegraphics[width=\h]{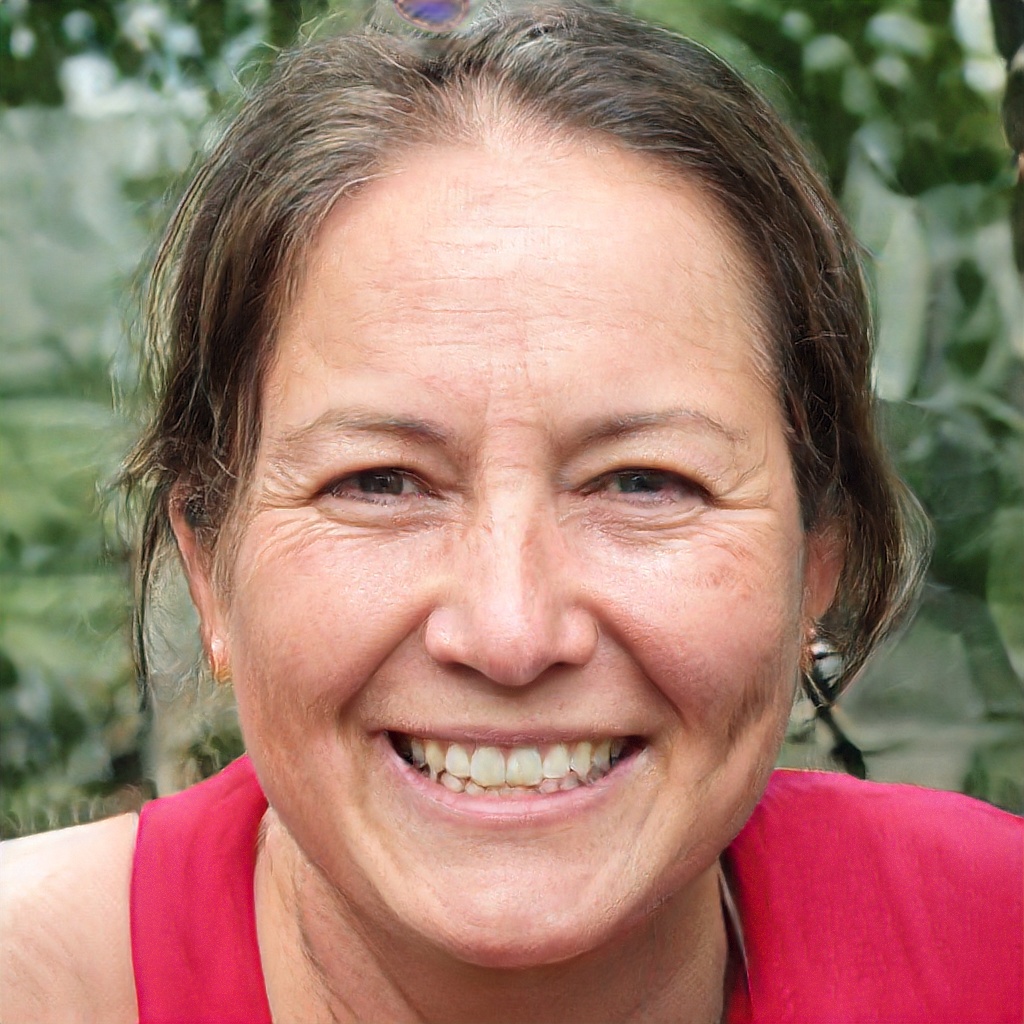}\hfill%
\includegraphics[width=\h]{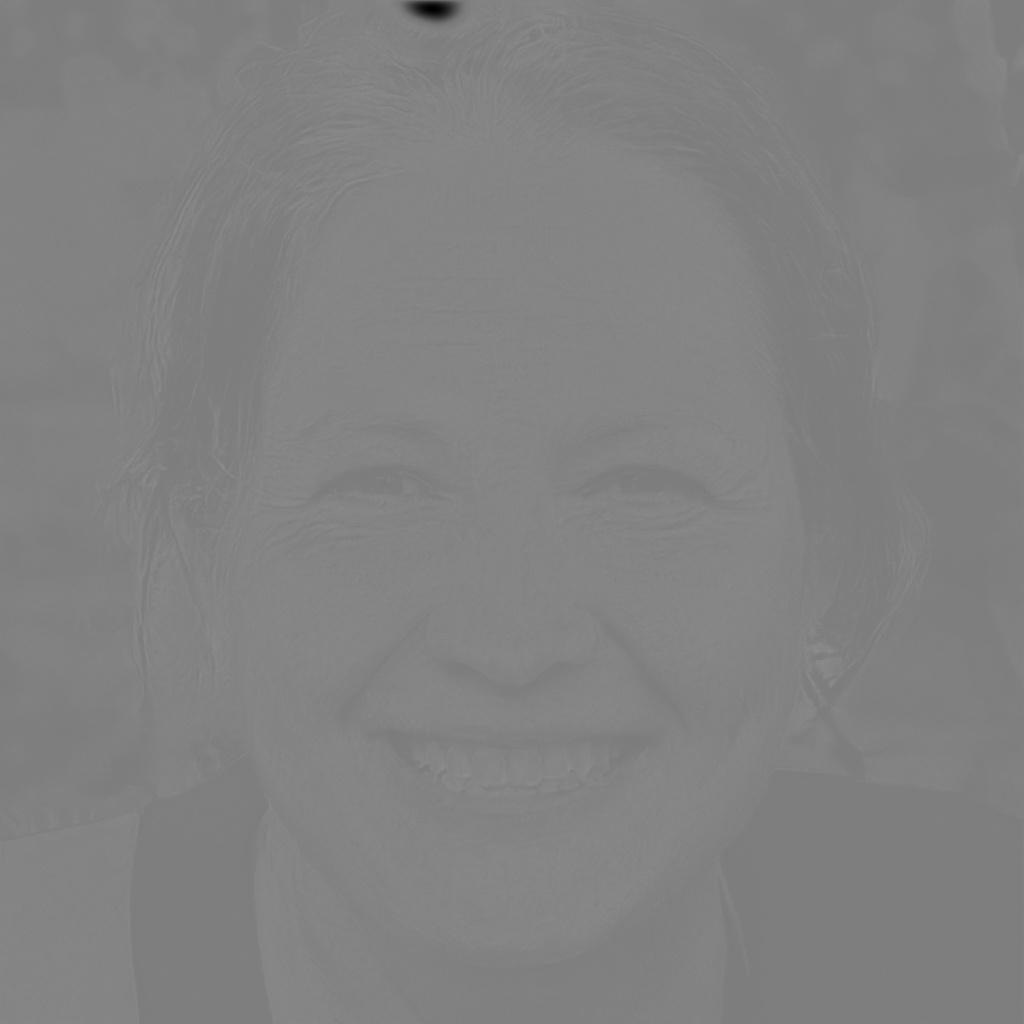}\hfill\hfill\hfill%
\includegraphics[width=\h,trim={32 66 100 66},clip]{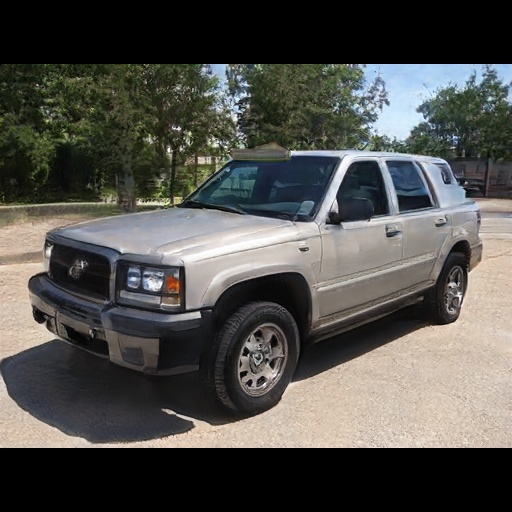}\hfill%
\includegraphics[width=\h,trim={16 33 50 33},clip]{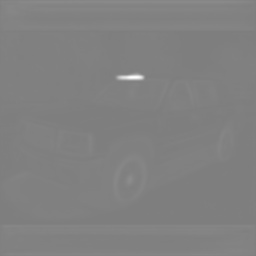}\hfill\hfill\hfill%
\includegraphics[width=\h]{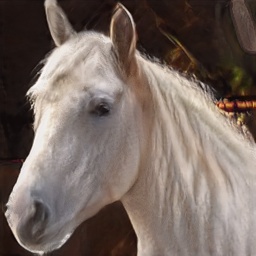}\hfill%
\includegraphics[width=\h]{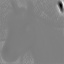}\\
\makebox[\h]{\hspace{-4.5mm}\raisebox{29.5mm}[0mm][0mm]{\begin{tikzpicture}\draw[red,thick] (0,0) circle (2.5mm);\end{tikzpicture}}}\hfill%
\makebox[\h]{\hspace{-4.5mm}\raisebox{29.5mm}[0mm][0mm]{\begin{tikzpicture}\draw[red,thick] (0,0) circle (2.5mm);\end{tikzpicture}}}\hfill\hfill\hfill%
\makebox[\h]{\hspace{5.4mm}\raisebox{22.5mm}[0mm][0mm]{\begin{tikzpicture}\draw[red,thick] (0,0) circle (3.5mm);\end{tikzpicture}}}\hfill%
\makebox[\h]{\hspace{5.4mm}\raisebox{22.5mm}[0mm][0mm]{\begin{tikzpicture}\draw[red,thick] (0,0) circle (3.5mm);\end{tikzpicture}}}\hfill\hfill\hfill%
\makebox[\h]{\hspace{24.5mm}\raisebox{25.5mm}[0mm][0mm]{\begin{tikzpicture}\draw[red,thick] (0,0) circle (4mm);\end{tikzpicture}}}\hfill%
\makebox[\h]{\hspace{24.5mm}\raisebox{25.5mm}[0mm][0mm]{\begin{tikzpicture}\draw[red,thick] (0,0) circle (4mm);\end{tikzpicture}}}\\\vspace{-1.5\baselineskip}
\caption{Instance normalization causes water droplet -like artifacts in StyleGAN images. These are not always obvious in the generated images, but if we look at the activations inside the generator network, the problem is always there, in all feature maps starting from the 64x64 resolution. It is a systemic problem that plagues all StyleGAN images.
}
\label{fig:droplets}
\end{figure*}
}

\newcommand{\figarch}{
\begin{figure*}[t]
\includegraphics[width=\linewidth,trim={0 81 0 74},clip]{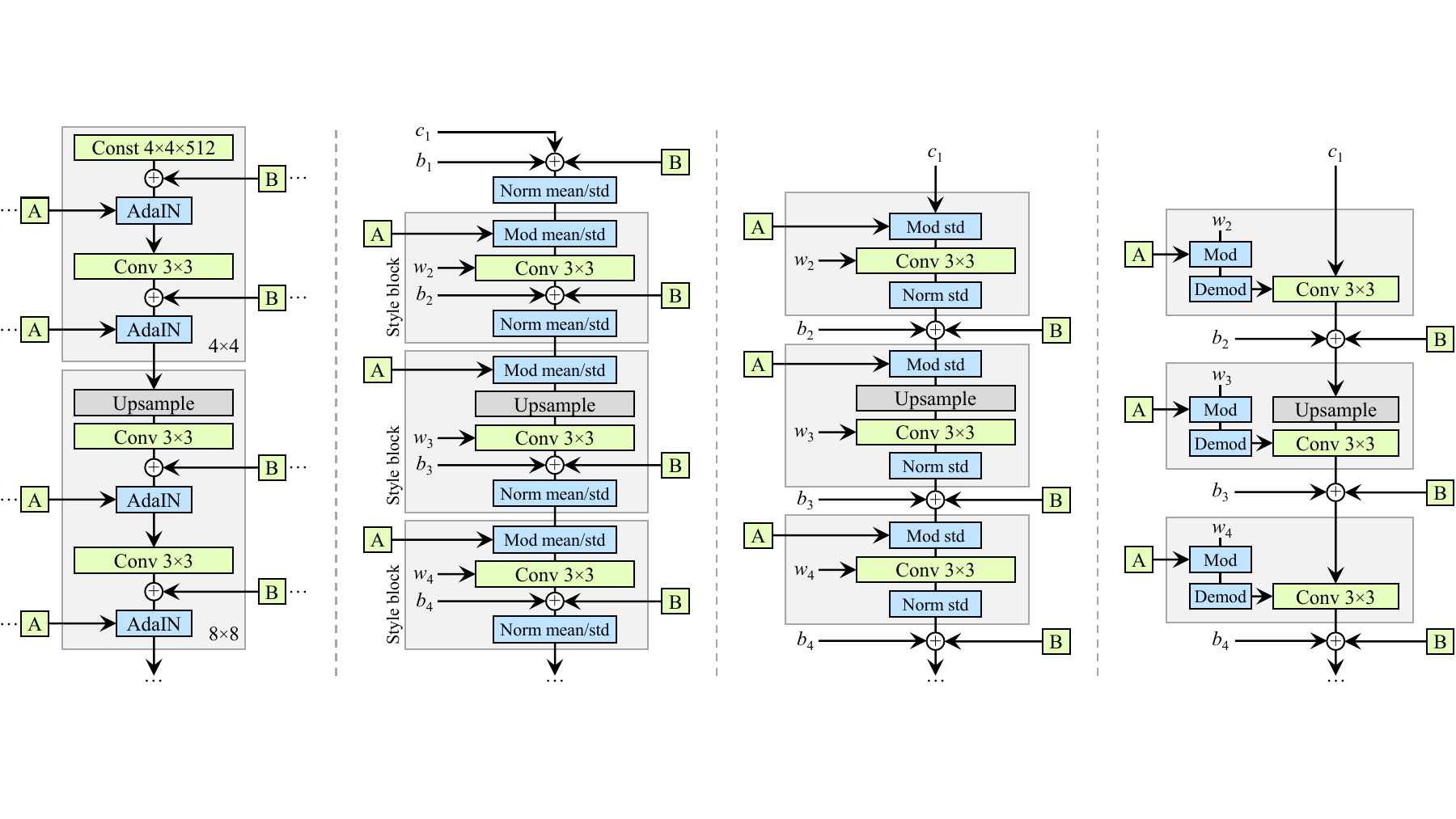}\\
\footnotesize
\makebox[0.216\linewidth]{(a) StyleGAN}\hfill%
\makebox[0.249\linewidth]{(b) StyleGAN (detailed)}\hfill%
\makebox[0.249\linewidth]{(c) Revised architecture}\hfill%
\makebox[0.231\linewidth]{(d) Weight demodulation}\\
\caption{
\setlength{\fboxsep}{1.5pt}
We redesign the architecture of the StyleGAN synthesis network. (a) The original StyleGAN, where \framebox{A} denotes a learned affine transform from $\WW$ that produces a style and \framebox{B} is a noise broadcast operation. (b) The same diagram with full detail. Here we have broken the AdaIN to explicit normalization followed by modulation, both operating on the mean and standard deviation per feature map. We have also annotated the learned weights ($w$), biases ($b)$, and constant input ($c$), and redrawn the gray boxes so that one style is active per box. The activation function (leaky ReLU) is always applied right after adding the bias. (c) We make several changes to the original architecture that are justified in the main text. We remove some redundant operations at the beginning, move the addition of $b$ and \framebox{B} to be outside active area of a style, and adjust only the standard deviation per feature map. (d) The revised architecture enables us to replace instance normalization with a ``demodulation'' operation, which we apply to the weights associated with each convolution layer.
}
\label{fig:arch}
\end{figure*}
}

\newcommand{\tabmain}{
\begin{table*}[t]
\centering
\newcolumntype{x}{>{\centering\arraybackslash\hspace{0pt}}p{11.9mm}}
\footnotesize{
\begin{tabular}{|l@{\hspace{1.5mm}}l|xxxx|xxxx|}
\hline
& \multirow{2}{*}{\textbf{Configuration}}
    & \multicolumn{4}{c|}{\textbf{FFHQ, 1024$\times$1024}}
    & \multicolumn{4}{c|}{\textbf{LSUN Car, 512$\times$384}}
\\
&
    & FID \FINAL{$\downarrow$}
    & \makebox[0mm][c]{Path length \FINAL{$\downarrow$}}
    & \makebox[0mm][c]{Precision \FINAL{$\uparrow$}}
    & Recall \FINAL{$\uparrow$}
    & FID \FINAL{$\downarrow$}
    & \makebox[0mm][c]{Path length \FINAL{$\downarrow$}}
    & \makebox[0mm][c]{Precision \FINAL{$\uparrow$}}
    & Recall \FINAL{$\uparrow$}
\\ \hline
\arch{a} & Baseline StyleGAN \cite{Karras2018}\ \                   %
    & 4.40          & 212.1         & {\bf 0.721}   & 0.399         %
    & 3.27          & 1484.5        & {\bf 0.701}   & 0.435         %
\\
\arch{b} & + Weight demodulation\ \                                 %
    & 4.39          & 175.4         & 0.702         & 0.425         %
    & 3.04          & \s862.4       & 0.685         & 0.488         %
\\
\arch{c} & + Lazy regularization\ \                                 %
    & 4.38          & 158.0         & 0.719         & 0.427         %
    & 2.83          & \s981.6       & 0.688         & 0.493         %
\\
\arch{d} & + Path length regularization\ \                          %
    & 4.34          & {\bf 122.5}   & 0.715         & 0.418         %
    & 3.43          & \s651.2       & 0.697         & 0.452         %
\\
\arch{e} & + No growing, new G \& D arch.\ \                        %
    & 3.31          & 124.5         & 0.705         & 0.449         %
    & 3.19          & \s471.2       & 0.690         & 0.454         %
\\ \hline
\arch{f} & + Large networks \FINAL{(StyleGAN2)}\ \                  %
    & {\bf 2.84}    & 145.0         & 0.689         & {\bf 0.492}   %
    & {\bf 2.32}    & {\bf \s415.5} & 0.678         & {\bf 0.514}   %
\\
& \FINAL{Config \arch{a} with large networks}\ \                    %
    & 3.98          & 199.2         & 0.716         & 0.422         %
    & --            & --            & --            & --            %
\\ \hline
\end{tabular}\vspace{2mm}}
\caption{
Main results.
For each training run, we selected the training snapshot with the lowest FID.
We computed each metric 10 times with different random seeds and report their average.
\FINAL{\emph{Path length}} corresponds to the PPL metric, computed based on path endpoints in $\WW$ \cite{Karras2018}\FINAL{, without the central crop used by Karras et al. \cite{Karras2018}.}
The \textsc{FFHQ} dataset contains 70k images, and the discriminator saw 25M images during training. 
For \textsc{LSUN Car} the numbers were 893k and 57M. 
\FINAL{$\uparrow$ indicates that higher is better, and $\downarrow$ that lower is better.}
}\vspace{-2mm}
\label{tab:main}
\end{table*}
}

\newcommand{\tabskips}{
\begin{table}[t]
\centering
\newcolumntype{x}{>{\centering\arraybackslash\hspace{0pt}}p{7mm}}
\footnotesize{
\begin{tabular}{|l|x@{\hspace{2.5mm}}x|x@{\hspace{2.5mm}}x|x@{\hspace{2.5mm}}x|}
\hline
\multirow{2}{*}{\textbf{FFHQ}}
    & \multicolumn{2}{c|}{D original}
    & \multicolumn{2}{c|}{D input skips}
    & \multicolumn{2}{c|}{D residual}
\\
    & FID & PPL
    & FID & PPL
    & FID & PPL
\\ \hline
G original
    & 4.32          & 265           %
    & 4.18          & 235           %
    & 3.58          & 269           %
\\
G output skips
    & 4.33          & 169           %
    & 3.77          & 127           %
    & {\bf 3.31}    & {\bf 125}     %
\\
G residual
    & 4.35          & 203           %
    & 3.96          & 229           %
    & 3.79          & 243           %
\\ \hline
\end{tabular}\vspace{1.5mm}\\
\begin{tabular}{|l|x@{\hspace{2.5mm}}x|x@{\hspace{2.5mm}}x|x@{\hspace{2.5mm}}x|}
\hline
\multirow{2}{*}{\textbf{LSUN Car}}
    & \multicolumn{2}{c|}{D original}
    & \multicolumn{2}{c|}{D input skips}
    & \multicolumn{2}{c|}{D residual}
\\
    & FID & PPL
    & FID & PPL
    & FID & PPL
\\ \hline
G original
    & 3.75          & 905           %
    & 3.23          & 758           %
    & 3.25          & 802           %
\\
G output skips
    & 3.77          & 544           %
    & 3.86          & {\bf 316}     %
    & 3.19          & 471           %
\\
G residual
    & 3.93          & 981           %
    & 3.40          & 667           %
    & {\bf 2.66}    & 645           %
\\ \hline
\end{tabular}\vspace{2mm}}
\caption{
Comparison of generator and discriminator architectures without progressive growing.
The combination of generator with output skips and residual discriminator corresponds to configuration~\arch{e} in the main result table.
}
\label{tab:skips}
\end{table}
}

\newcommand{\tablsun}{
\begin{table}[t]
\centering
\newcolumntype{x}{>{\centering\arraybackslash\hspace{0pt}}p{7.2mm}}
\footnotesize{
\begin{tabular}{|l|c|x@{\hspace{2.5mm}}x|x@{\hspace{2.5mm}}x|}
\hline
\multirow{2}{*}{\textbf{Dataset}} & \multirow{2}{*}{\textbf{Resolution}}
    & \multicolumn{2}{c|}{\textbf{StyleGAN} (\arch{a})}
    & \multicolumn{2}{c|}{\textbf{StyleGAN2} (\arch{f})}
\\
&
    & FID & PPL
    & FID & PPL
\\ \hline
\textsc{LSUN Car} & 512$\times$384
    & 3.27          & 1485          %
    & {\bf 2.32}    & {\bf 416}     %
\\
\textsc{LSUN Cat} & 256$\times$256
    & 8.53          & \s924         %
    & {\bf 6.93}    & {\bf 439}     %
\\
\textsc{LSUN Church} & 256$\times$256
    & 4.21          & \s742         %
    & {\bf 3.86}    & {\bf 342}     %
\\
\textsc{LSUN Horse} & 256$\times$256
    & 3.83          & 1405          %
    & {\bf 3.43}    & {\bf 338}     %
\\ \hline
\end{tabular}\vspace{2mm}}
\caption{
Improvement in LSUN datasets measured using FID and PPL.
We trained \textsc{Car} for 57M images, \textsc{Cat} for 88M, \textsc{Church} for 48M, and \textsc{Horse} for 100M images.
}
\label{tab:lsun}
\end{table}
}

\newcommand{\figpplimages}{
\begin{figure}[t]
\renewcommand{\h}{0.48\linewidth}
\includegraphics[width=\h]{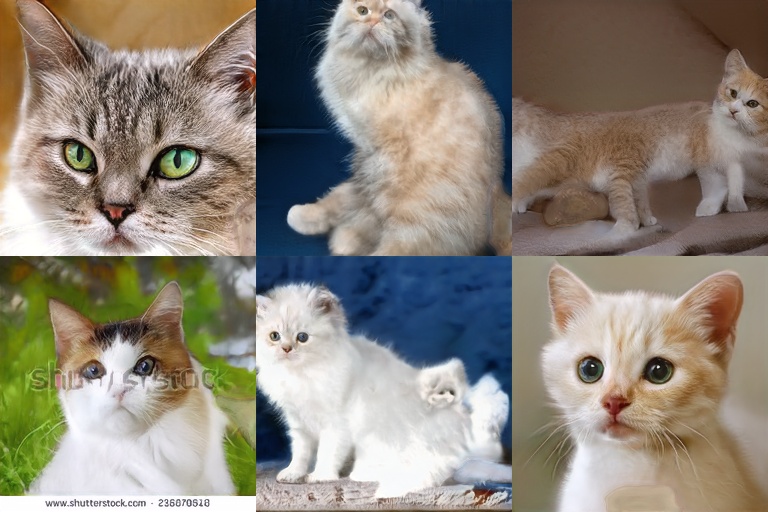}\hfill%
\includegraphics[width=\h]{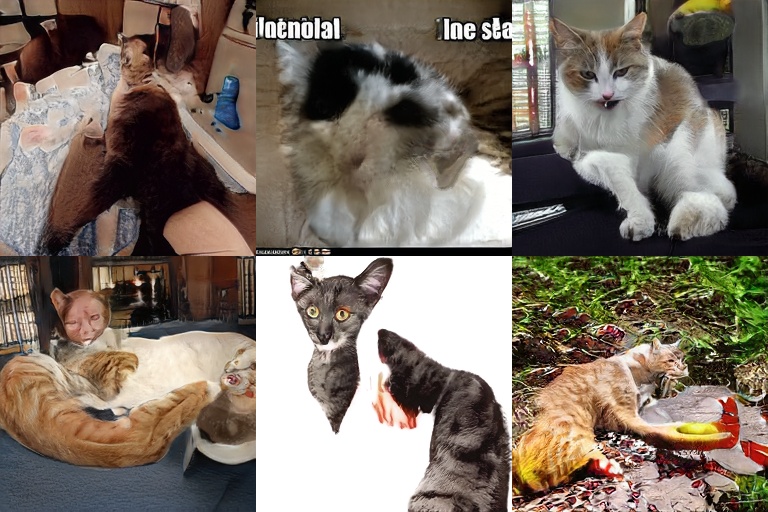}\vspace*{-1mm}\\
\makebox[\h][c]{\footnotesize (a) Low PPL scores}\hfill%
\makebox[\h][c]{\footnotesize (b) High PPL scores}\\\vspace{-0.5\baselineskip}
\caption{
Connection between perceptual path length and image quality using baseline StyleGAN \FINAL{(config \arch{a}) with \textsc{LSUN Cat}}.
(a) Random examples with low PPL ($\le 10^\mathrm{th}$ percentile).
(b) Examples with high PPL ($\ge 90^\mathrm{th}$ percentile).
There is a clear correlation between PPL scores and semantic consistency of the images.
}
\label{fig:pplimages}
\end{figure}
}

\newcommand{\figpplhistograms}{
\begin{figure}[t]
\renewcommand{\h}{0.49\linewidth}
\includegraphics[width=\h]{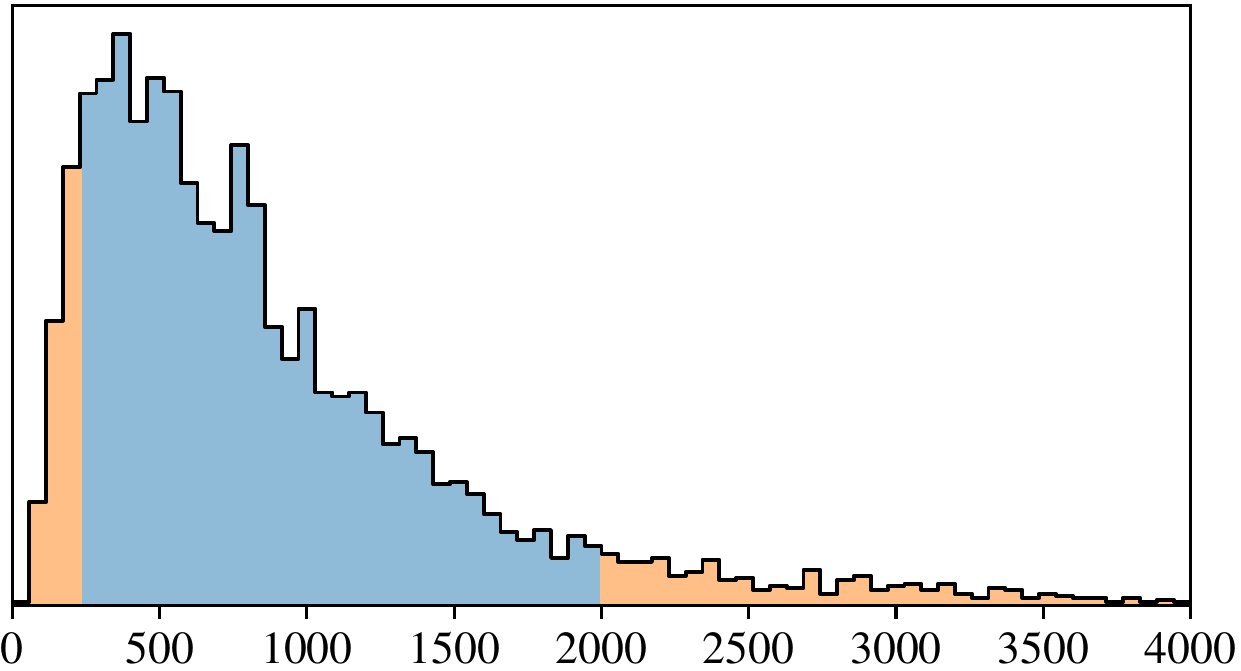}\hfill%
\includegraphics[width=\h]{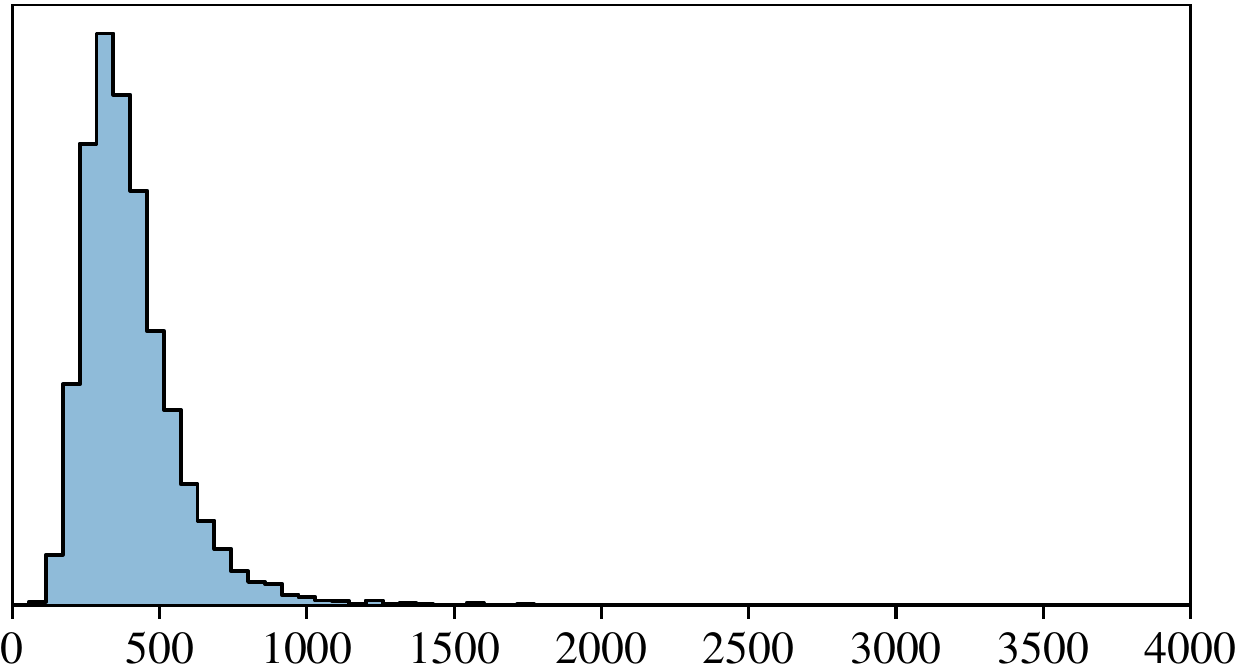}\vspace*{-1mm}\\
\makebox[\h][c]{\footnotesize (a) StyleGAN (config \arch{a})}\hfill%
\makebox[\h][c]{\footnotesize (b) StyleGAN2 (config \arch{f})}\\\vspace{-0.5\baselineskip}
\caption{
(a) Distribution of PPL scores of individual images generated using baseline StyleGAN \FINAL{(config \arch{a}) with \textsc{LSUN Cat} (FID\,=\,8.53, PPL\,=\,924)}.
The percentile ranges corresponding to Figure~\ref{fig:pplimages} are highlighted in orange.
(b) StyleGAN2 (config \arch{f}) improves the PPL distribution considerably (showing a snapshot with the same FID\,=\,8.53, PPL\,=\,387).
}
\label{fig:pplhistograms}
\end{figure}
}

\newcommand{\figGDarch}{
\begin{figure}[t]\vspace*{-0.75mm}
\includegraphics[width=\linewidth,trim={132 84 134 47},clip]{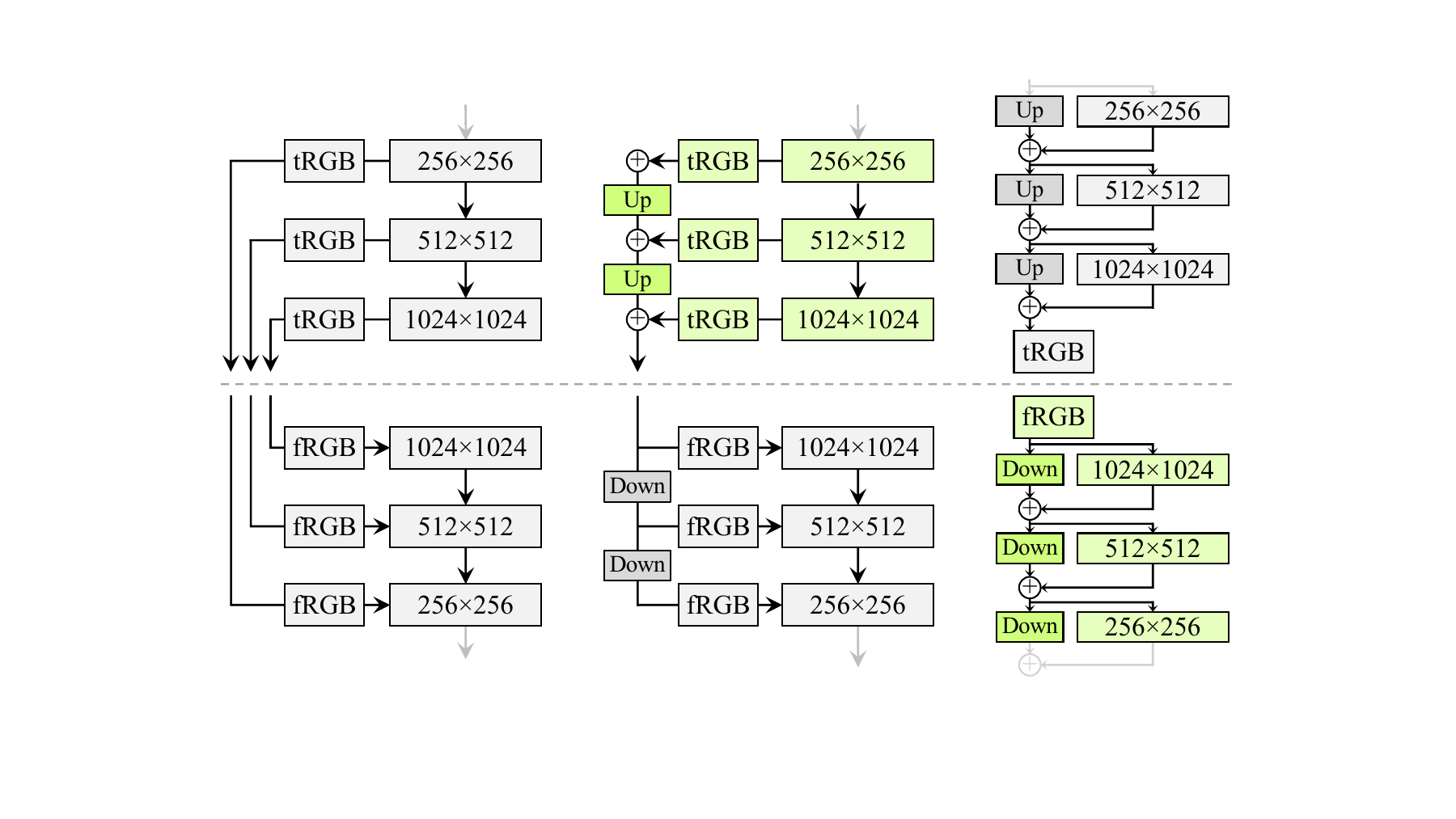}\\
\footnotesize
\makebox[0.33\linewidth]{(a) MSG-GAN}\hfill%
\makebox[0.36\linewidth]{(b) Input/output skips}\hfill%
\makebox[0.22\linewidth]{(c) Residual nets}\\
\caption{
\setlength{\fboxsep}{1.5pt}
Three generator (above the dashed line) and discriminator architectures.
\framebox{Up} and \framebox{Down} denote bilinear up and downsampling, respectively.
In residual networks these also include 1$\times$1 convolutions to adjust the number of feature maps.
\framebox{tRGB} and \framebox{fRGB} convert between RGB and high-dimensional per-pixel data.
Architectures used in configs \arch{e} and \arch{f} are shown in green.
}
\label{fig:GDarch}
\end{figure}
}

\newcommand{\figVIN}{
\begin{figure}[t]
\renewcommand{\h}{0.3305\linewidth}
\includegraphics[width=\h]{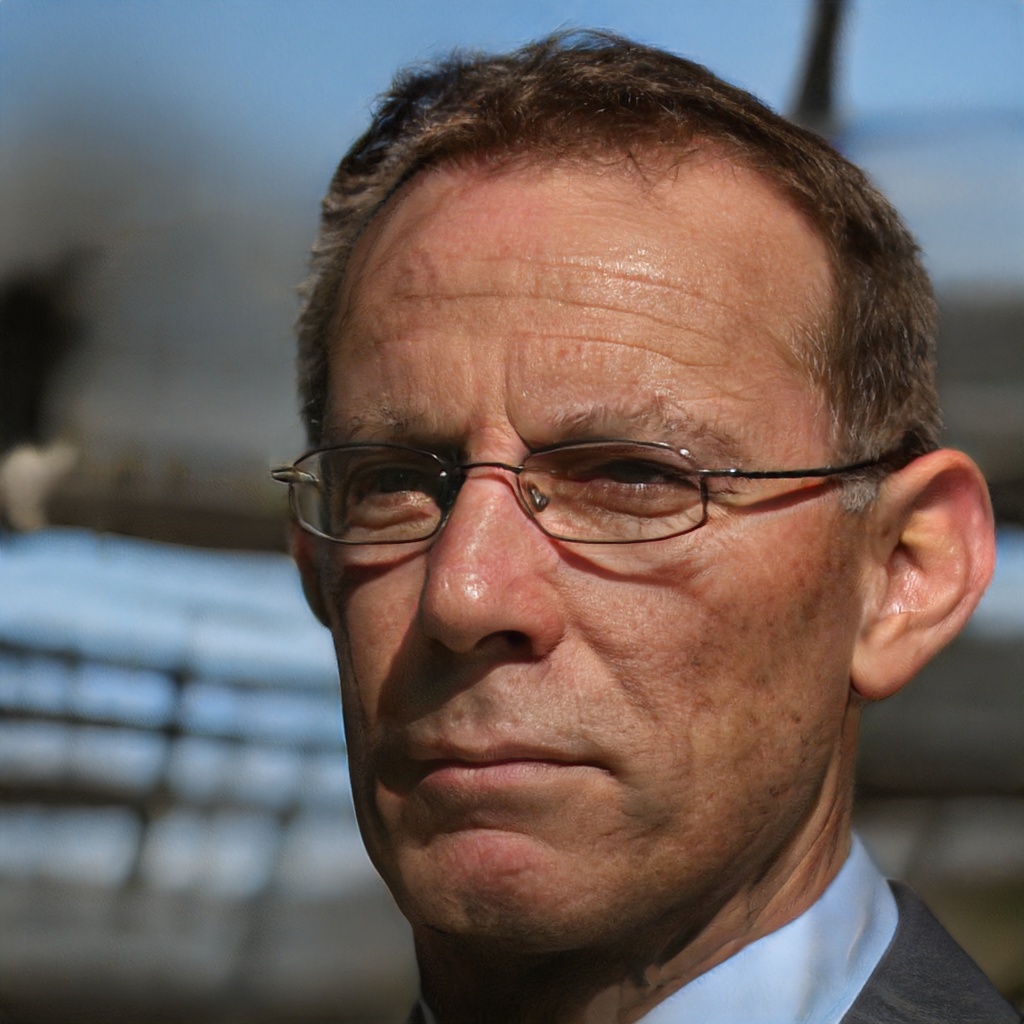}\hfill%
\includegraphics[width=\h]{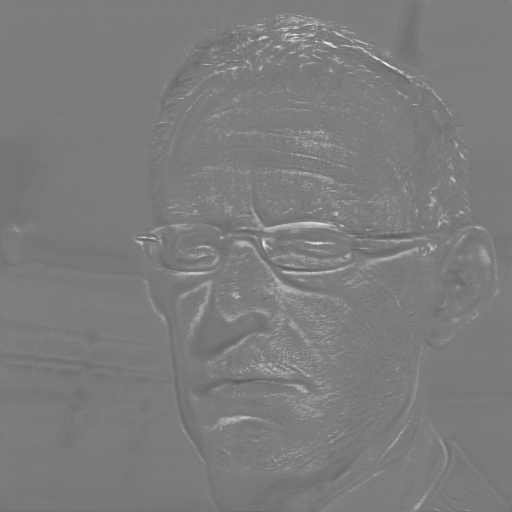}\hfill%
\includegraphics[width=\h]{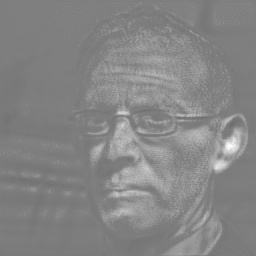}\\
\includegraphics[width=\h]{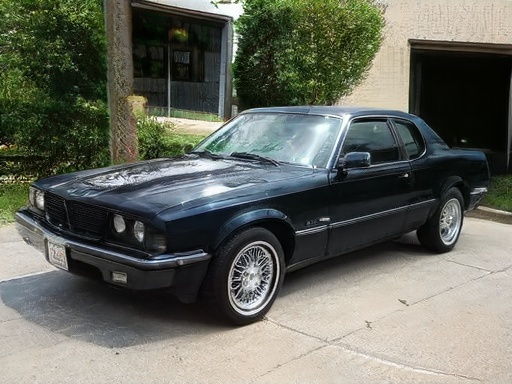}\hfill%
\includegraphics[width=\h]{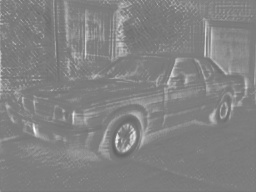}\hfill%
\includegraphics[width=\h]{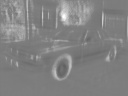}\\
\includegraphics[width=\h]{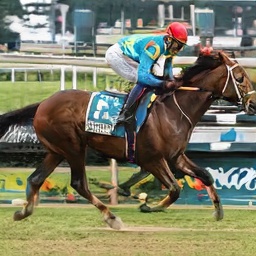}\hfill%
\includegraphics[width=\h]{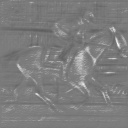}\hfill%
\includegraphics[width=\h]{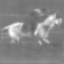}\\\vspace{-0.5\baselineskip}
\caption{Replacing normalization with demodulation removes the characteristic artifacts from images and activations.
}
\label{fig:VIN}
\end{figure}
}

\newcommand{\figpgartifacts}{
\begin{figure}[t]
\renewcommand{\h}{0.3305\linewidth}
\includegraphics[width=\h]{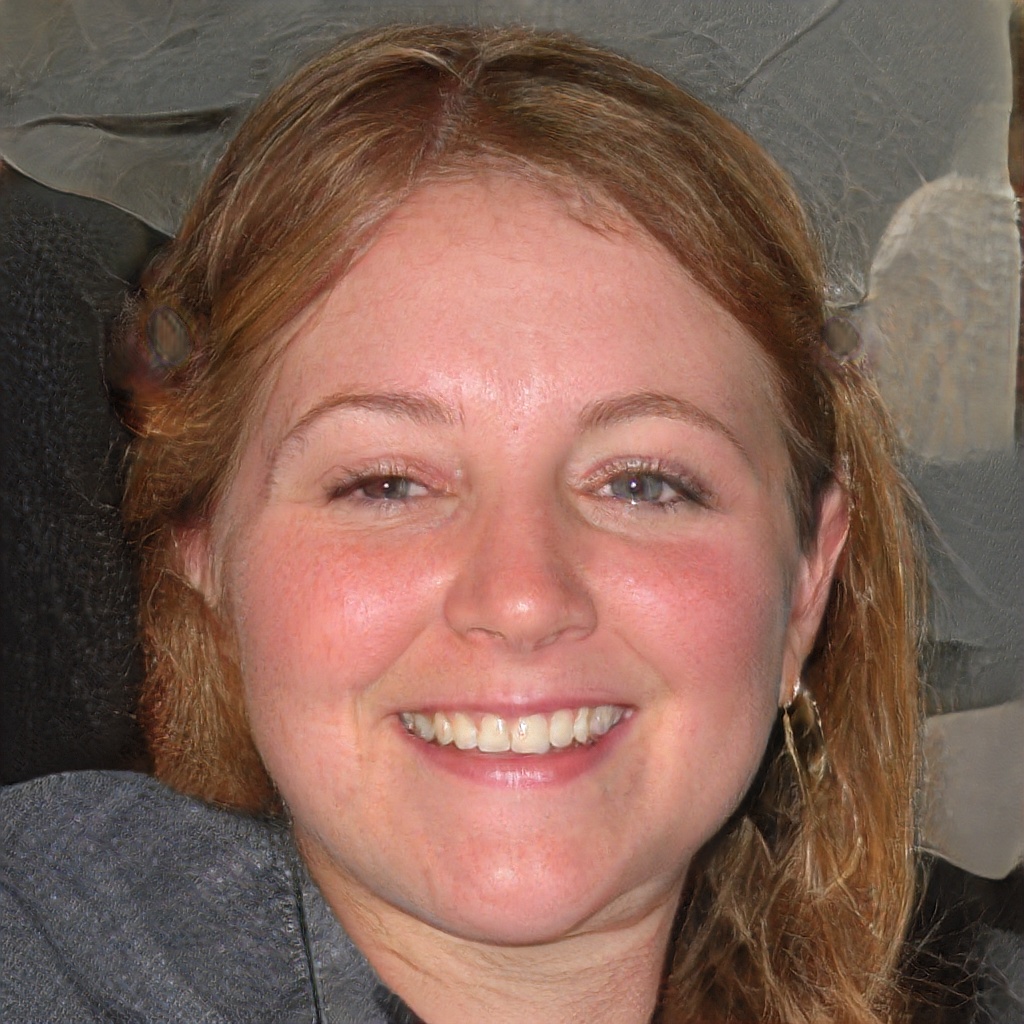}\hfill%
\includegraphics[width=\h]{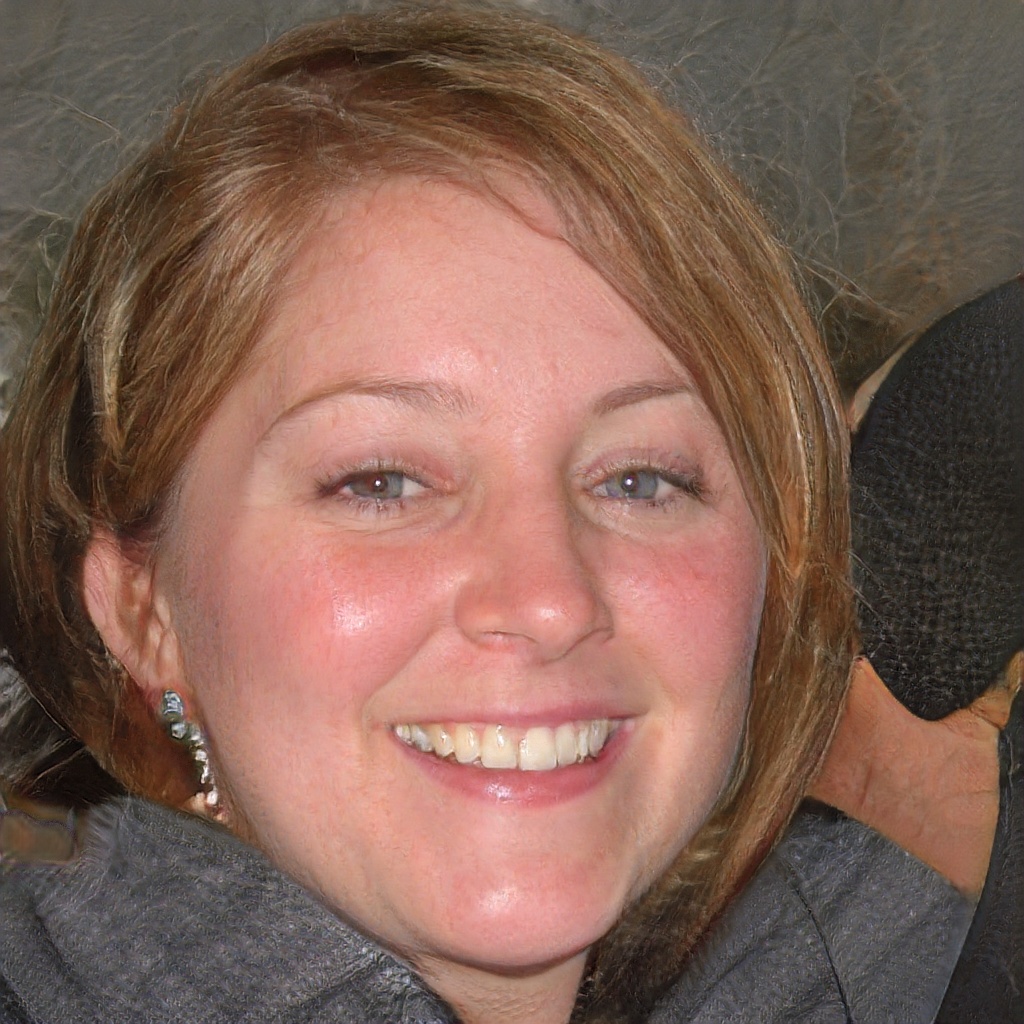}\hfill%
\includegraphics[width=\h]{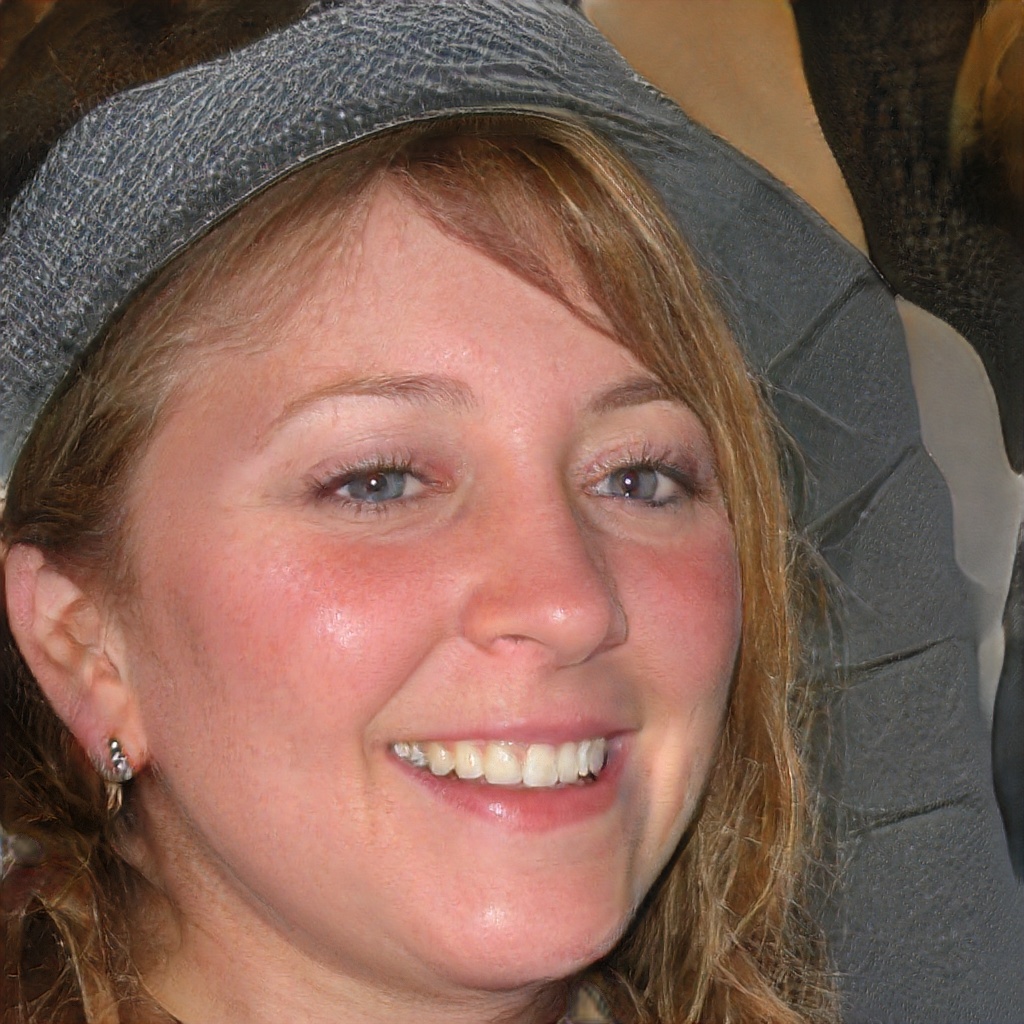}\\
\includegraphics[width=\h,trim={360 210 360 670},clip]{figures/artifacts/teeth4.\ext}\hfill%
\includegraphics[width=\h,trim={365 200 355 680},clip]{figures/artifacts/teeth3.\ext}\hfill%
\includegraphics[width=\h,trim={368 180 352 700},clip]{figures/artifacts/teeth1.\ext}\\
\makebox[\h]{\hspace{0mm}\raisebox{6mm}[0mm][0mm]{\begin{tikzpicture}\draw[blue] (0,0) -- (0,2.1);\end{tikzpicture}}}\hfill%
\makebox[\h]{\hspace{0.4mm}\raisebox{6mm}[0mm][0mm]{\begin{tikzpicture}\draw[blue] (0,0) -- (0,2.1);\end{tikzpicture}}}\hfill%
\makebox[\h]{\hspace{0.5mm}\raisebox{6mm}[0mm][0mm]{\begin{tikzpicture}\draw[blue] (0,0) -- (0,2.1);\end{tikzpicture}}}\vspace{-1.5\baselineskip}\\
\caption{Progressive growing leads to ``phase'' artifacts. In this example the teeth do not follow the pose but stay aligned to the camera, as indicated by the blue line.
}
\label{fig:pgartifacts}
\end{figure}
}

\newcommand{\figresolutionusage}{
\begin{figure}[t]
\renewcommand{\h}{0.495\linewidth}
\includegraphics[width=\h]{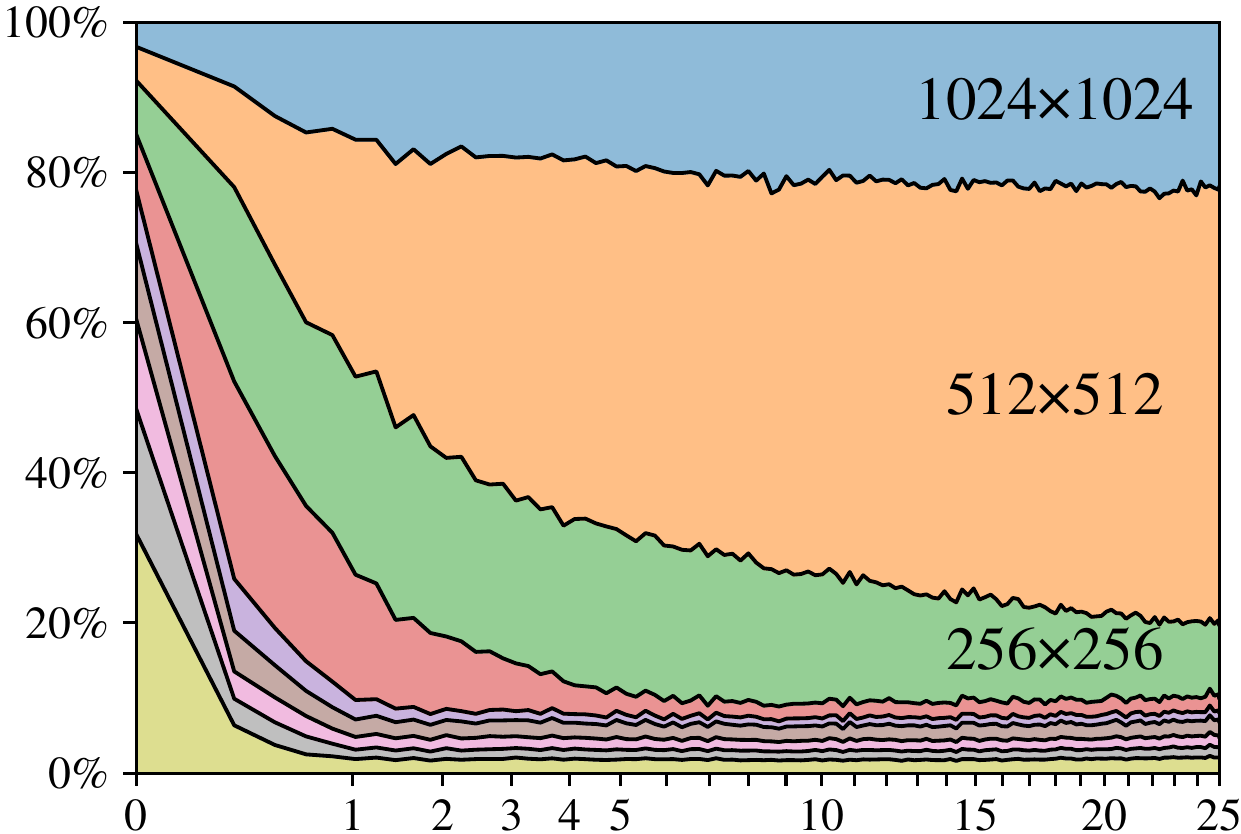}\hfill
\includegraphics[width=\h]{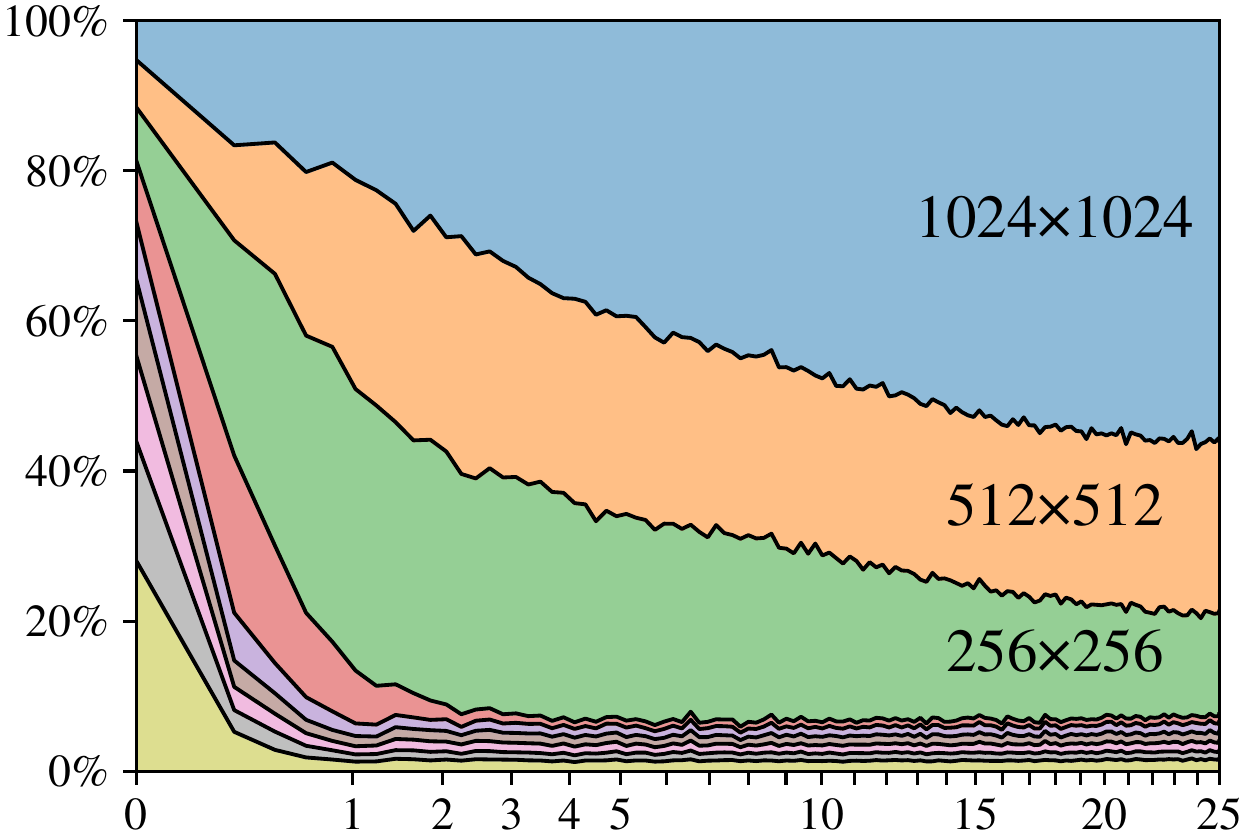}\\
\footnotesize%
\makebox[\h][c]{(a) StyleGAN-sized (config \arch{e})}\hfill%
\makebox[\h][c]{(b) Large networks (config \arch{f})}\\
\caption{
Contribution of each resolution to the output of the generator as a function of training time.
The vertical axis shows a breakdown of the relative standard deviations of different resolutions, and the horizontal axis corresponds to training progress, measured in millions of training images shown to the discriminator. 
We can see that in the beginning the network focuses on low-resolution images and progressively shifts its focus on larger resolutions as training progresses.
In (a) the generator basically outputs a $512^2$ image with some minor sharpening for $1024^2$, while in (b) the larger network focuses more on the high-resolution details.
}
\label{fig:resolutionusage}
\end{figure}
}

\newcommand{\figprojhist}{
\begin{figure}[t]
\renewcommand{\h}{0.49\linewidth}
\includegraphics[width=\h]{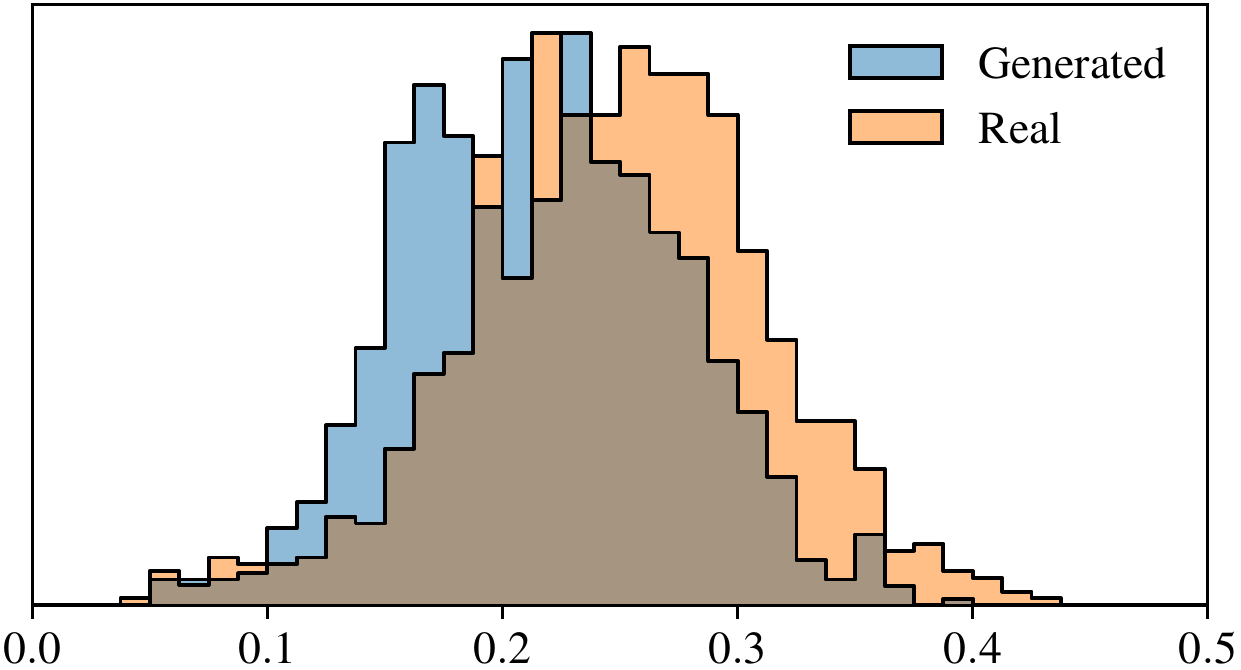}\hfill%
\includegraphics[width=\h]{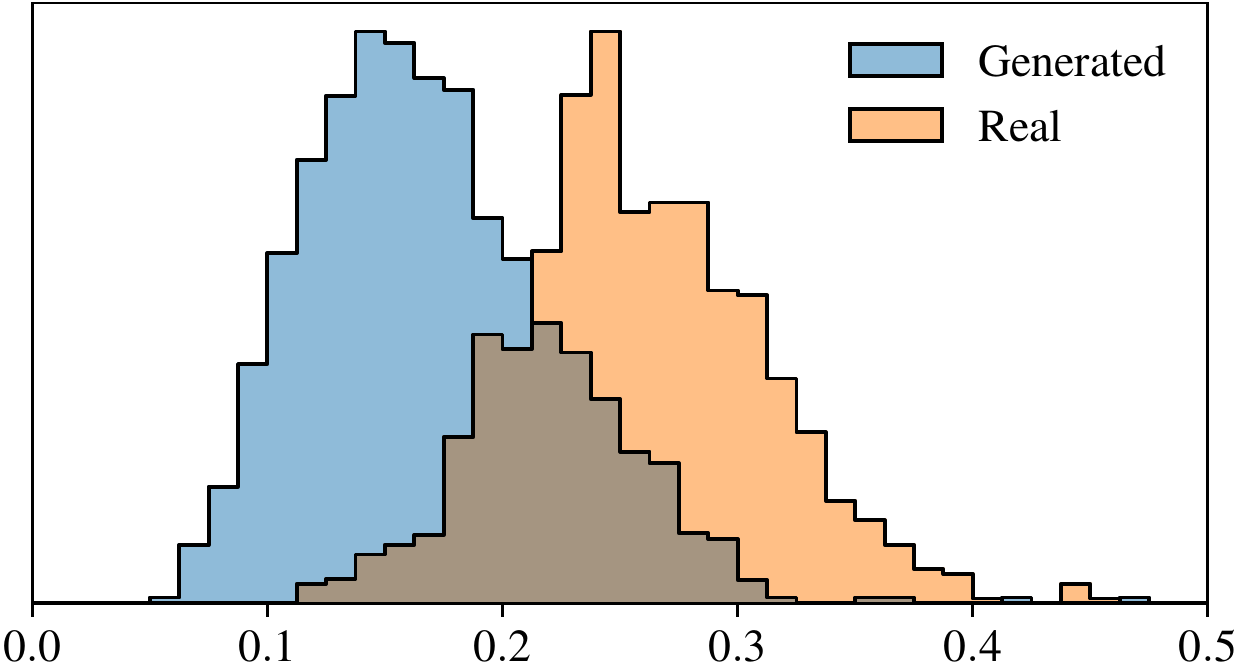}\vspace*{-1.5mm}\\
\makebox[\h][c]{\footnotesize \textsc{LSUN Car}, StyleGAN}\hfill%
\makebox[\h][c]{\footnotesize \textsc{FFHQ}, StyleGAN}\vspace*{1.5mm}\\
\includegraphics[width=\h]{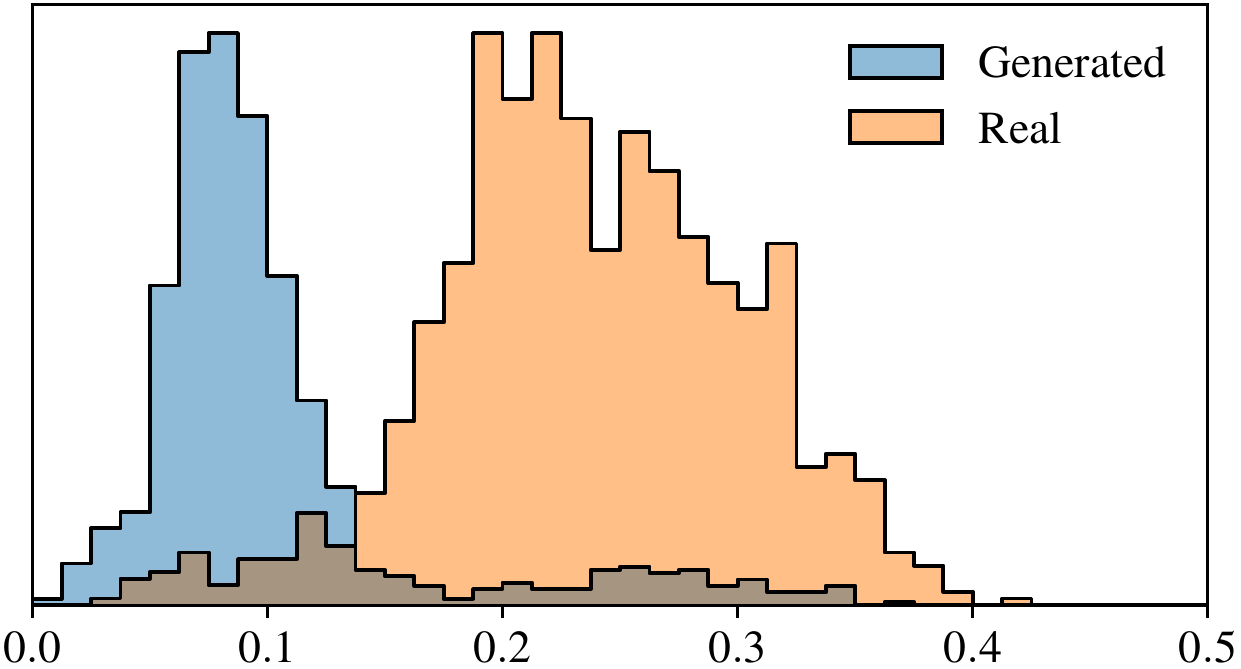}\hfill%
\includegraphics[width=\h]{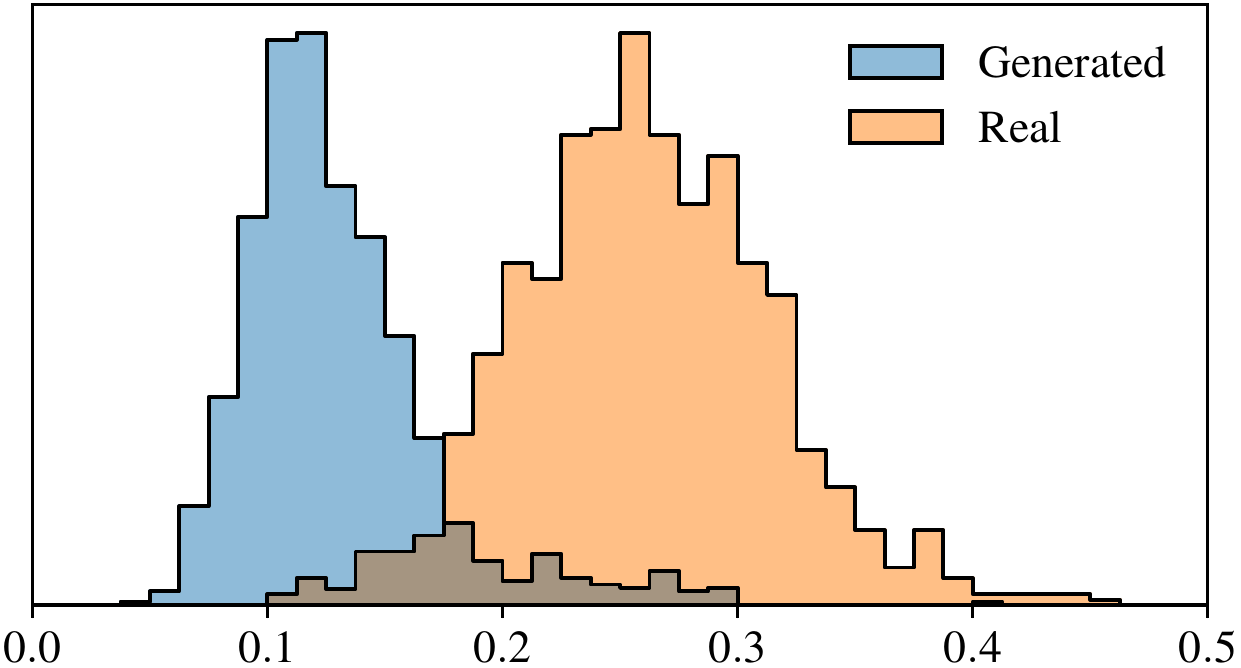}\vspace*{-1.5mm}\\
\makebox[\h][c]{\footnotesize \textsc{LSUN Car}, StyleGAN2}\hfill%
\makebox[\h][c]{\footnotesize \textsc{FFHQ}, StyleGAN2}\vspace*{-2mm}\\
\caption{\label{fig:projhist}%
LPIPS distance histograms between original and projected images for generated (blue) and real images (orange).
Despite the higher image quality of our improved generator, it is much easier to project the generated images into its latent space $\WW$.
The same projection method was used in all cases.
}
\end{figure}
}

\newcommand{\figprojimgcarhorizontal}{
\begin{figure*}[t]
\centering
\includegraphics[width=0.11\linewidth]{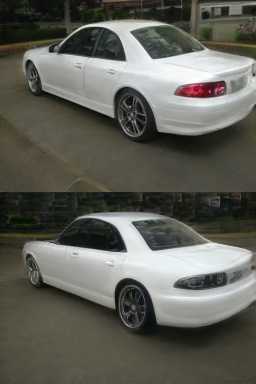}%
\includegraphics[width=0.11\linewidth]{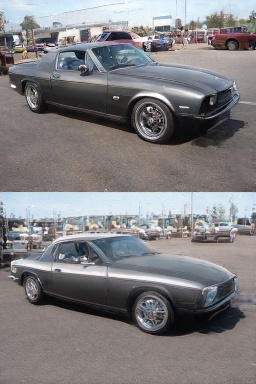}%
\includegraphics[width=0.11\linewidth]{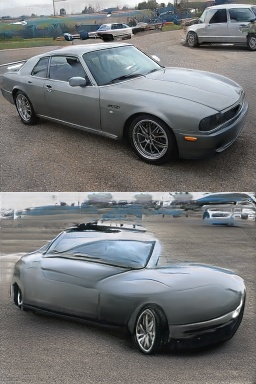}\hfill
\includegraphics[width=0.11\linewidth]{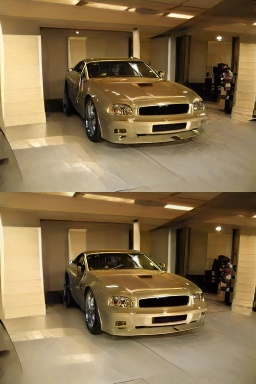}%
\includegraphics[width=0.11\linewidth]{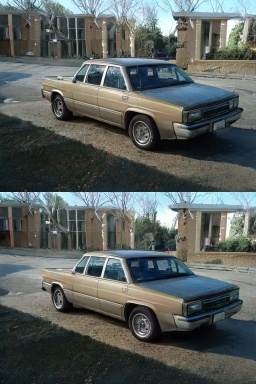}%
\includegraphics[width=0.11\linewidth]{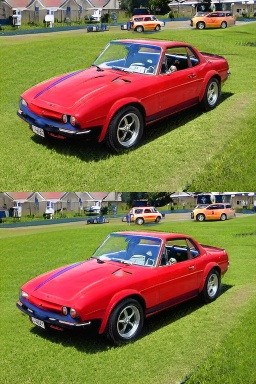}\hfill
\includegraphics[width=0.11\linewidth]{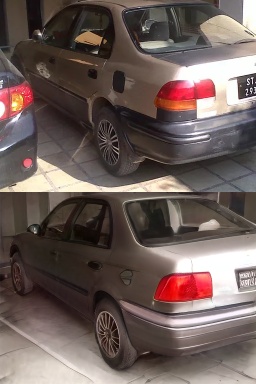}%
\includegraphics[width=0.11\linewidth]{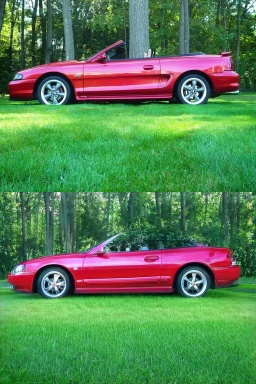}%
\includegraphics[width=0.11\linewidth]{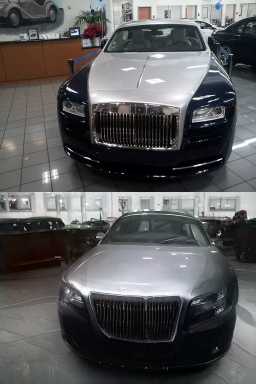}\vspace*{-1mm}\\
\makebox[0.33\linewidth]{\footnotesize \FINAL{StyleGAN --- generated images}}\hfill
\makebox[0.33\linewidth]{\footnotesize \FINAL{StyleGAN2 --- generated images}}\hfill
\makebox[0.33\linewidth]{\footnotesize \FINAL{StyleGAN2 --- real images}}\vspace*{1mm}\\
\caption{\label{fig:projimgcar}%
Example images and their projected and re-synthesized counterparts. 
For each configuration, top row shows the target images and bottom row shows the synthesis of the corresponding projected latent vector and noise inputs.
With the baseline StyleGAN, projection often finds a reasonably close match for generated images, but especially the backgrounds differ from the originals. 
The images generated using StyleGAN2 can be projected almost perfectly back into generator inputs, while projected real images (from the training set) show clear differences to the originals, as expected.
All tests were done using the same projection method and hyperparameters.
}
\end{figure*}
}

\newcommand{\figartifactsapp}{
\renewcommand{\h}{0.195\linewidth}
\begin{figure*}[t]
\includegraphics[width=\h]{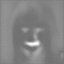}\hfill%
\includegraphics[width=\h]{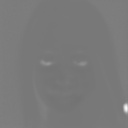}\hfill%
\includegraphics[width=\h]{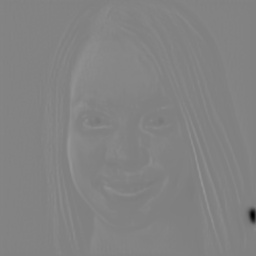}\hfill%
\includegraphics[width=\h]{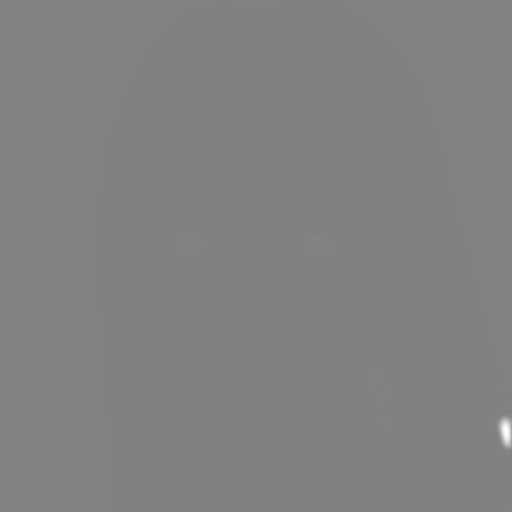}\hfill\hfill%
\includegraphics[width=\h]{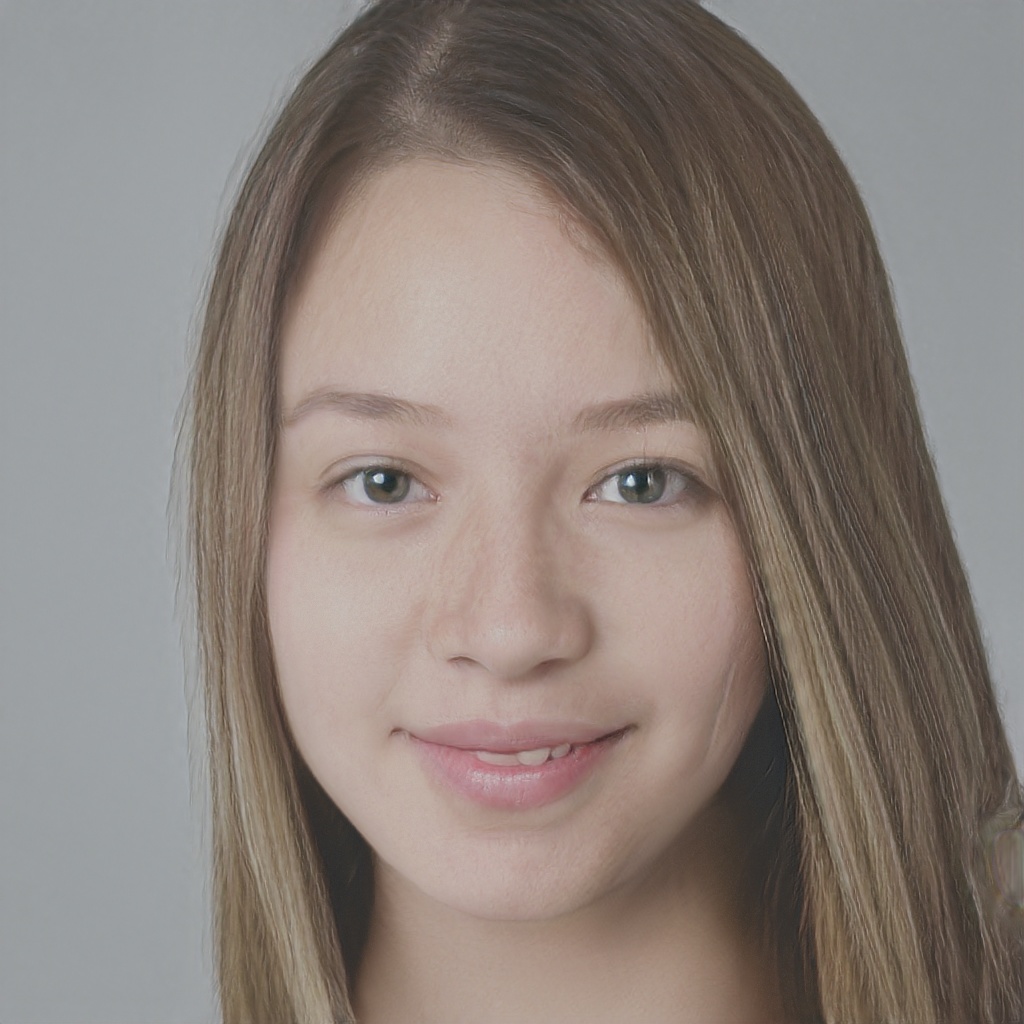}\vspace*{.5mm}\\
\includegraphics[width=\h]{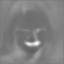}\hfill%
\includegraphics[width=\h]{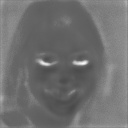}\hfill%
\includegraphics[width=\h]{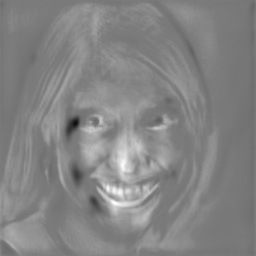}\hfill%
\includegraphics[width=\h]{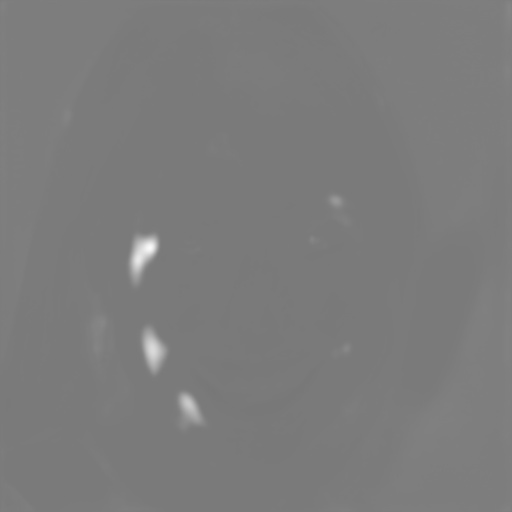}\hfill\hfill%
\includegraphics[width=\h]{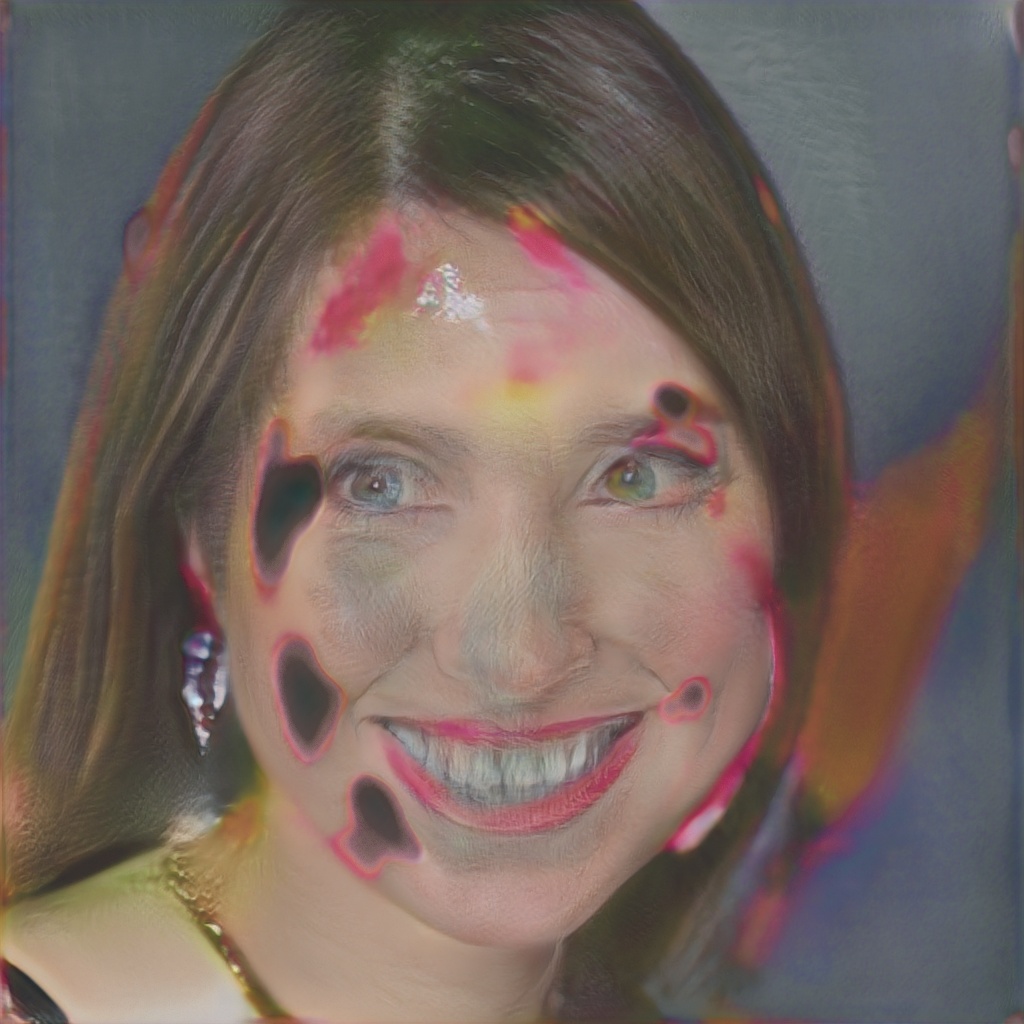}\\
\footnotesize 
\makebox[\h]{Feature map $64^2$}\hfill%
\makebox[\h]{Feature map $128^2$}\hfill%
\makebox[\h]{Feature map $256^2$}\hfill%
\makebox[\h]{Feature map $512^2$}\hfill\hfill%
\makebox[\h]{Generated image}\\
\caption{
An example of the importance of the droplet artifact in StyleGAN generator.
We compare two generated images, one successful and one severely corrupted.
The corresponding feature maps were normalized to the viewable dynamic range using instance normalization.
For the top image, the droplet artifact starts forming in $64^2$ resolution, is clearly visible in $128^2$, and increasingly dominates the feature maps in higher resolutions.
For the bottom image, $64^2$ is qualitatively similar to the top row, but the droplet does not materialize in $128^2$.
Consequently, the facial features are stronger in the normalized feature map.
This leads to an overshoot in $256^2$, followed by multiple spurious droplets forming in subsequent resolutions.
Based on our experience, it is rare that the droplet is missing from StyleGAN images, and indeed the generator fully relies on its existence. 
}
\label{fig:artifactsapp}
\end{figure*}
}

\newcommand{\figcarscomparison}{
\begin{figure*}[t]
\renewcommand{\h}{\linewidth}
\renewcommand{\hh}{3mm}
\vspace{\hh}%
\includegraphics[width=\h]{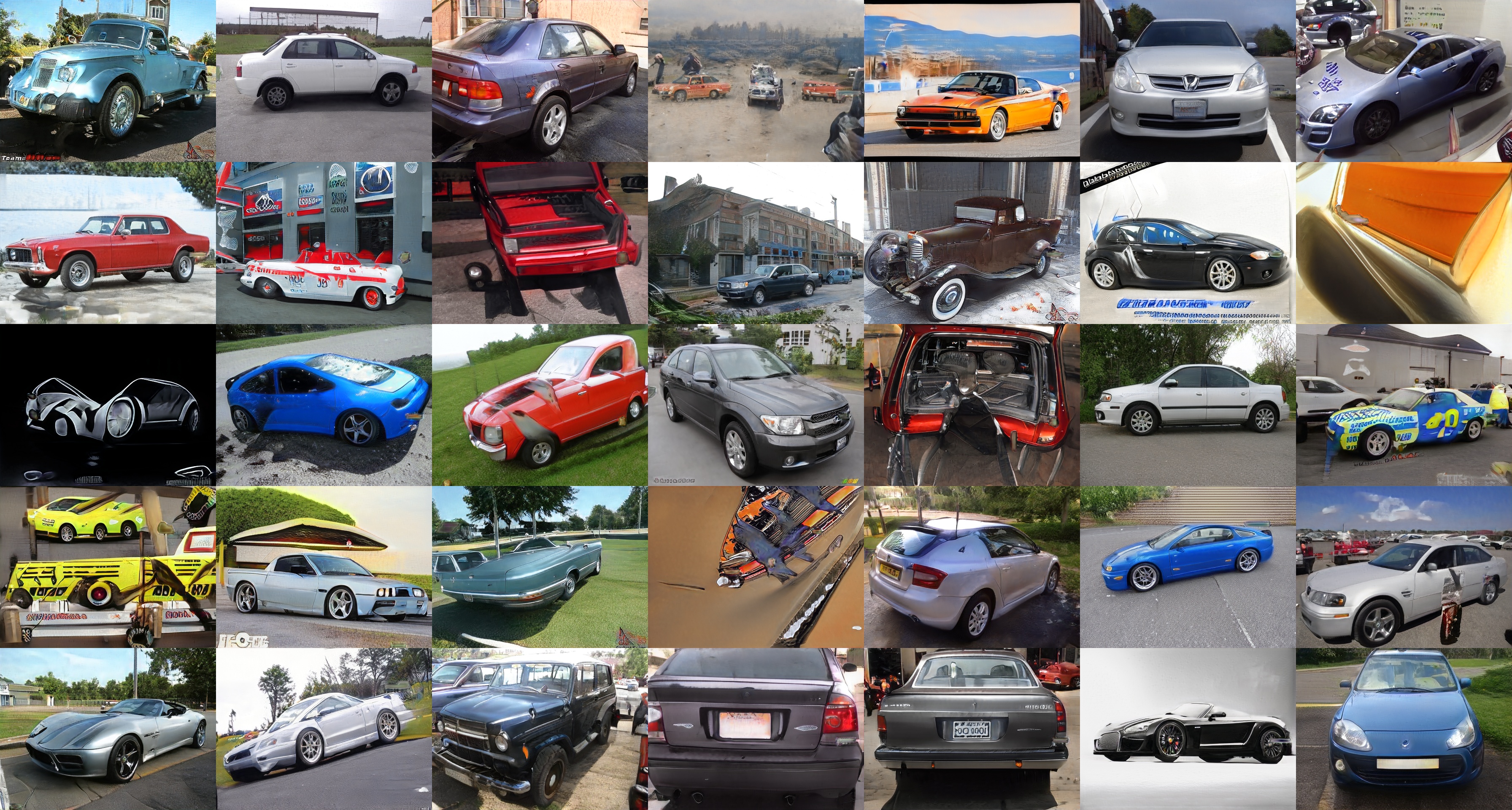}\\
\makebox[\h]{Model 1: FID\,=\,3.27, P\,=\,0.70, R\,=\,0.44, PPL\,=\,1485}\\\\
\includegraphics[width=\h]{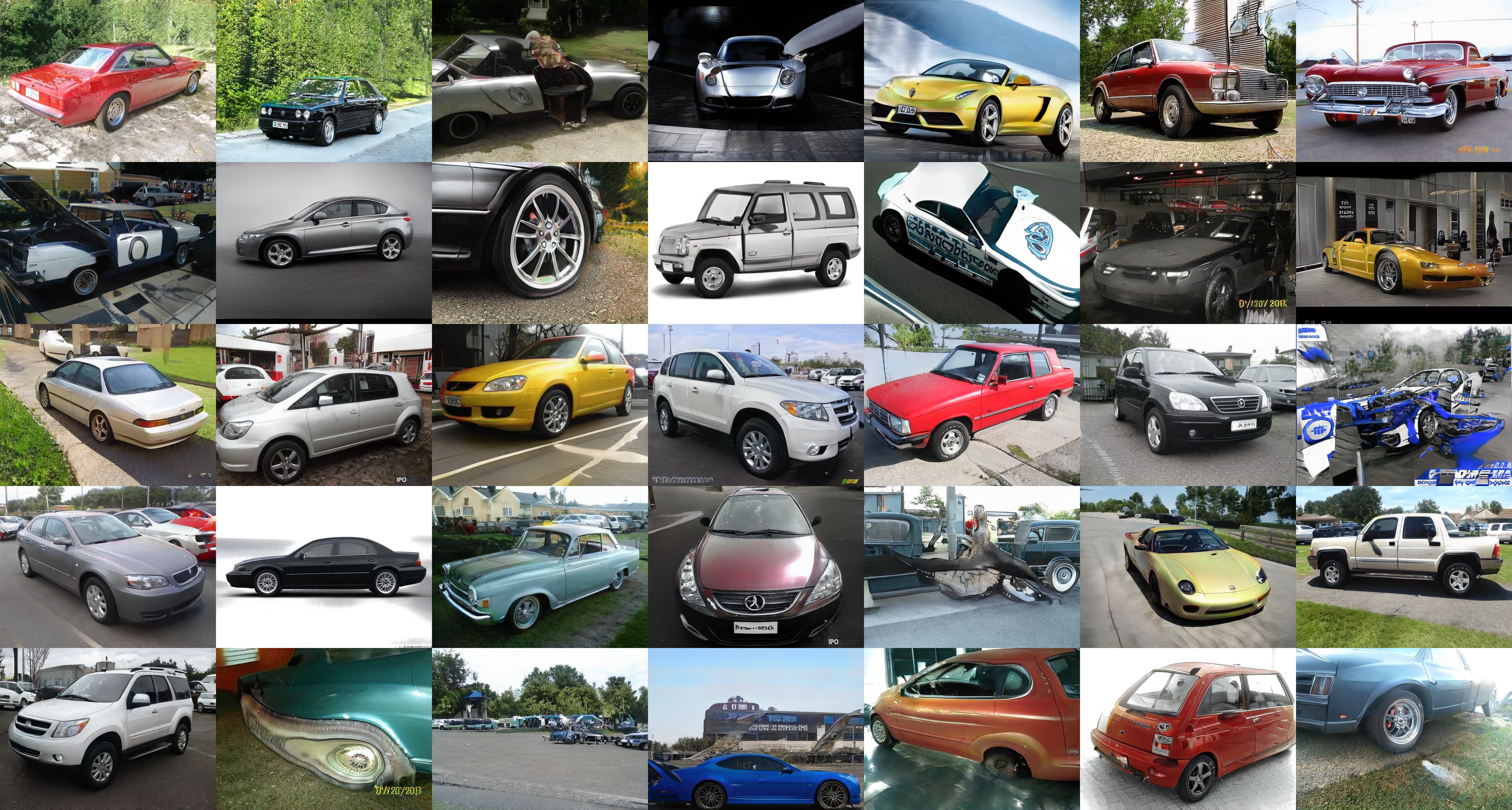}\\
\makebox[\h]{Model 2: FID\,=\,3.27, P\,=\,0.67, R\,=\,0.48, PPL\,=\,437}\\
\caption{%
Uncurated examples from two generative models trained on \textsc{LSUN Car} without truncation.
FID, precision, and recall are similar for models~1 and~2, even though the latter produces car-shaped objects more often.
Perceptual path length (PPL) indicates a clear preference for model~2. 
Model 1 corresponds to configuration~\arch{a} in \reftablsun{}, and model~2 is an early training snapshot of configuration~\arch{f}.
}
\vspace{\hh}
\label{fig:carscomparison}
\end{figure*}
}

\newcommand{\figcatscomparison}{
\begin{figure*}[t]
\renewcommand{\h}{\linewidth}
\renewcommand{\hh}{9mm}
\vspace{\hh}%
\includegraphics[width=\h]{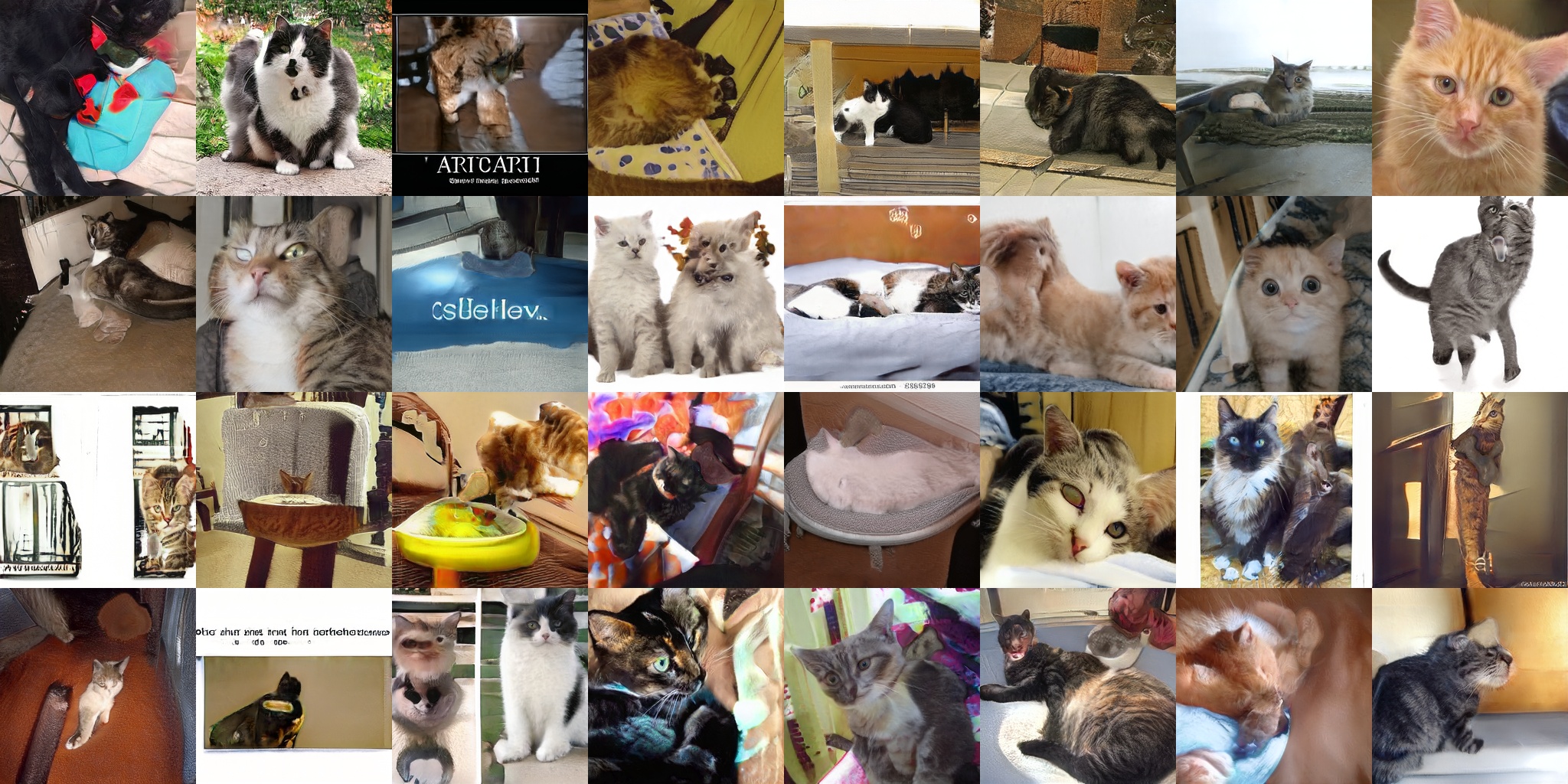}\\
\makebox[\h]{Model 1: FID\,=\,8.53, P\,=\,0.64, R\,=\,0.28, PPL\,=\,924}\\\\
\includegraphics[width=\h]{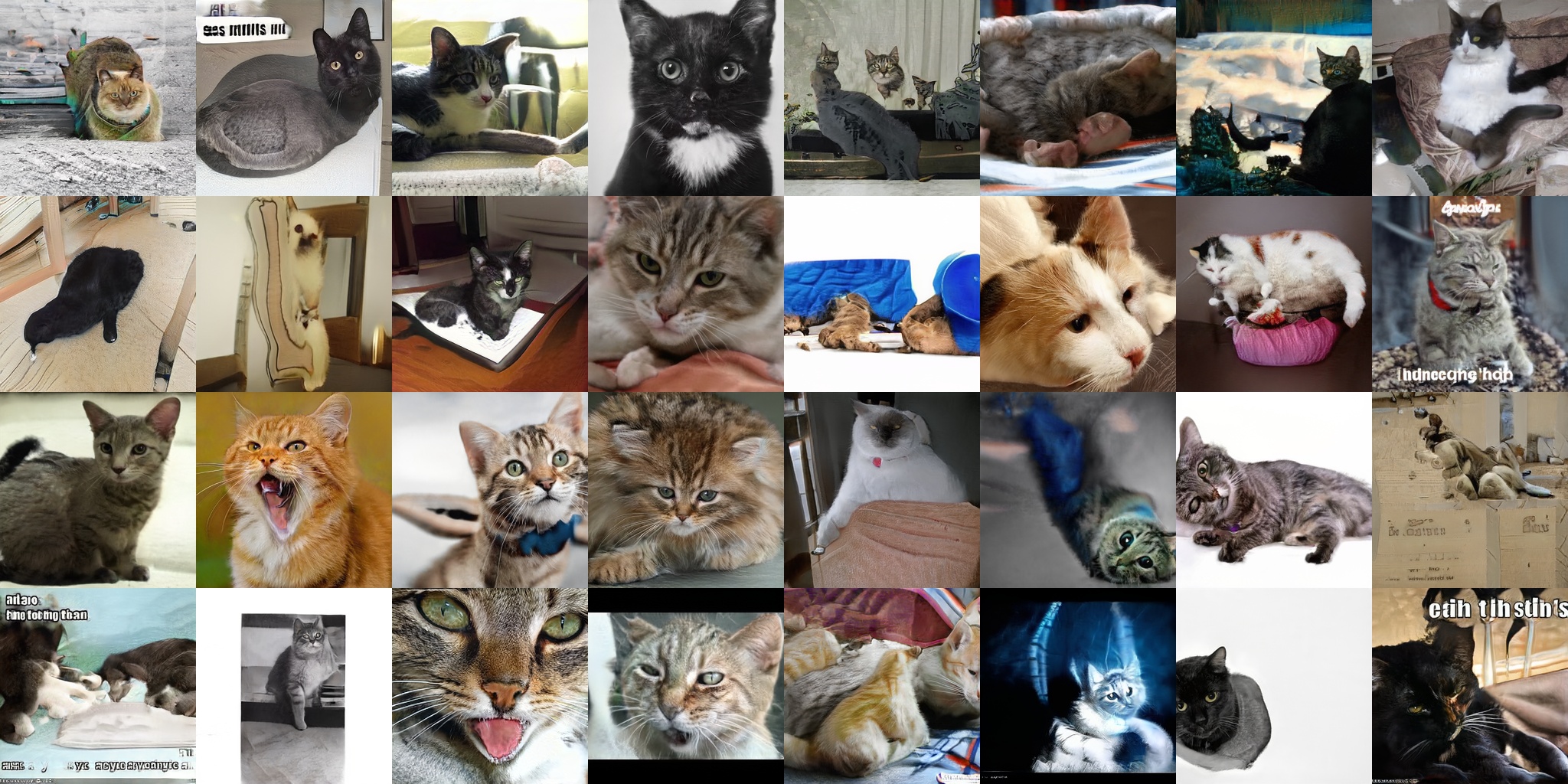}\\
\makebox[\h]{Model 2: FID\,=\,8.53, P\,=\,0.62, R\,=\,0.29, PPL\,=\,387}\\
\caption{%
Uncurated examples from two generative models trained on \textsc{LSUN Cat} without truncation.
FID, precision, and recall are similar for models~1 and~2, even though the latter produces cat-shaped objects more often.
Perceptual path length (PPL) indicates a clear preference for model~2. 
Model 1 corresponds to configuration~\arch{a} in \reftablsun{}, and model~2 is an early training snapshot of configuration~\arch{f}.
}
\vspace{\hh}
\label{fig:catscomparison}
\end{figure*}
}

\newcommand{\figffhqcherrypick}{
\begin{figure*}[t]
\renewcommand{\h}{0.499\linewidth}
\renewcommand{\hh}{20mm}
\vspace{\hh}%
\includegraphics[width=\h]{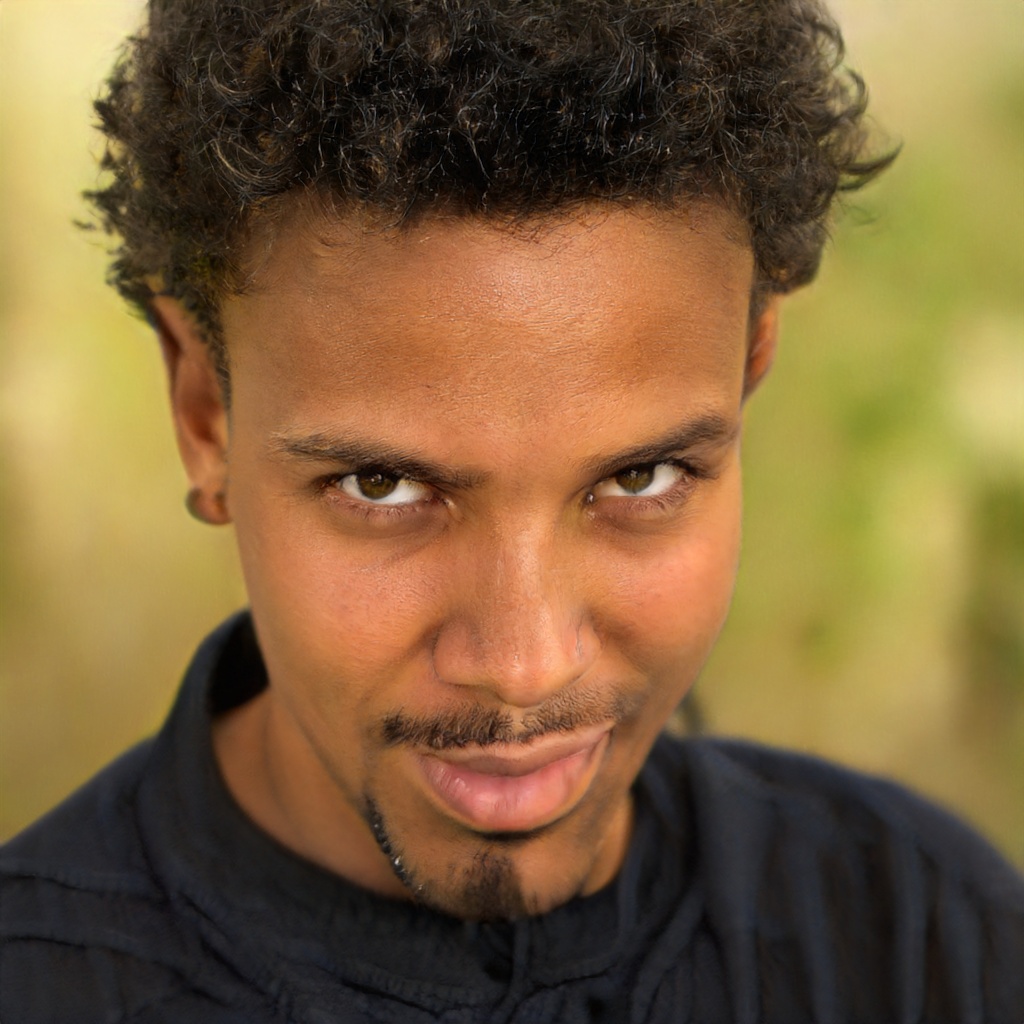}\hfill%
\includegraphics[width=\h]{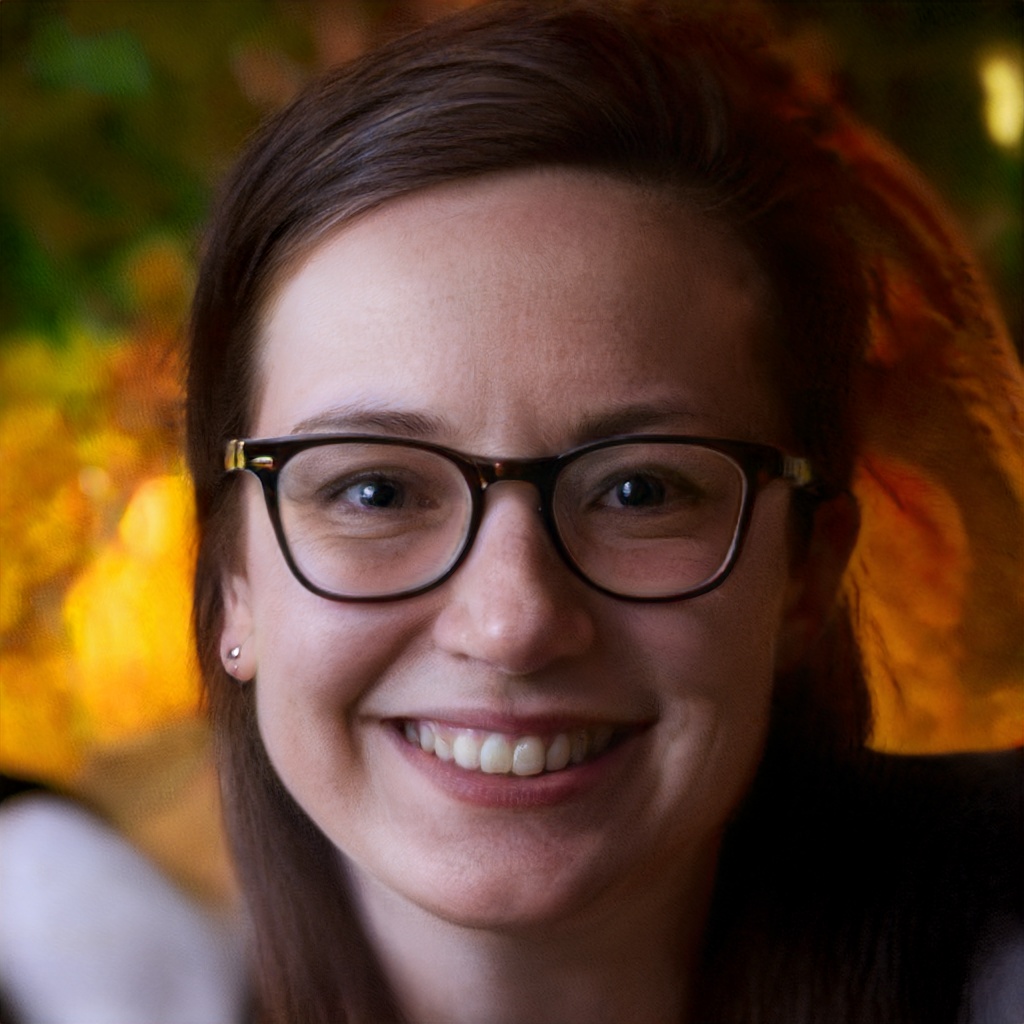}\\
\includegraphics[width=\h]{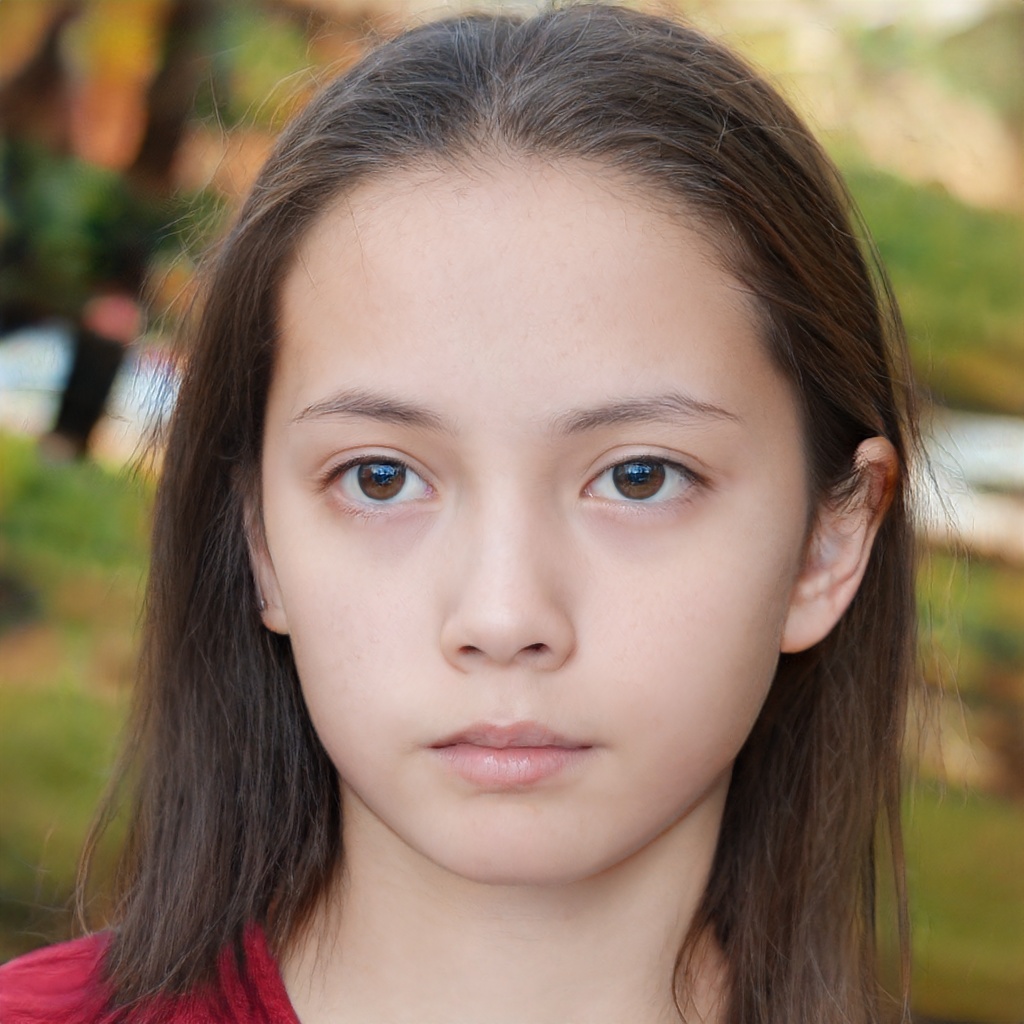}\hfill%
\includegraphics[width=\h]{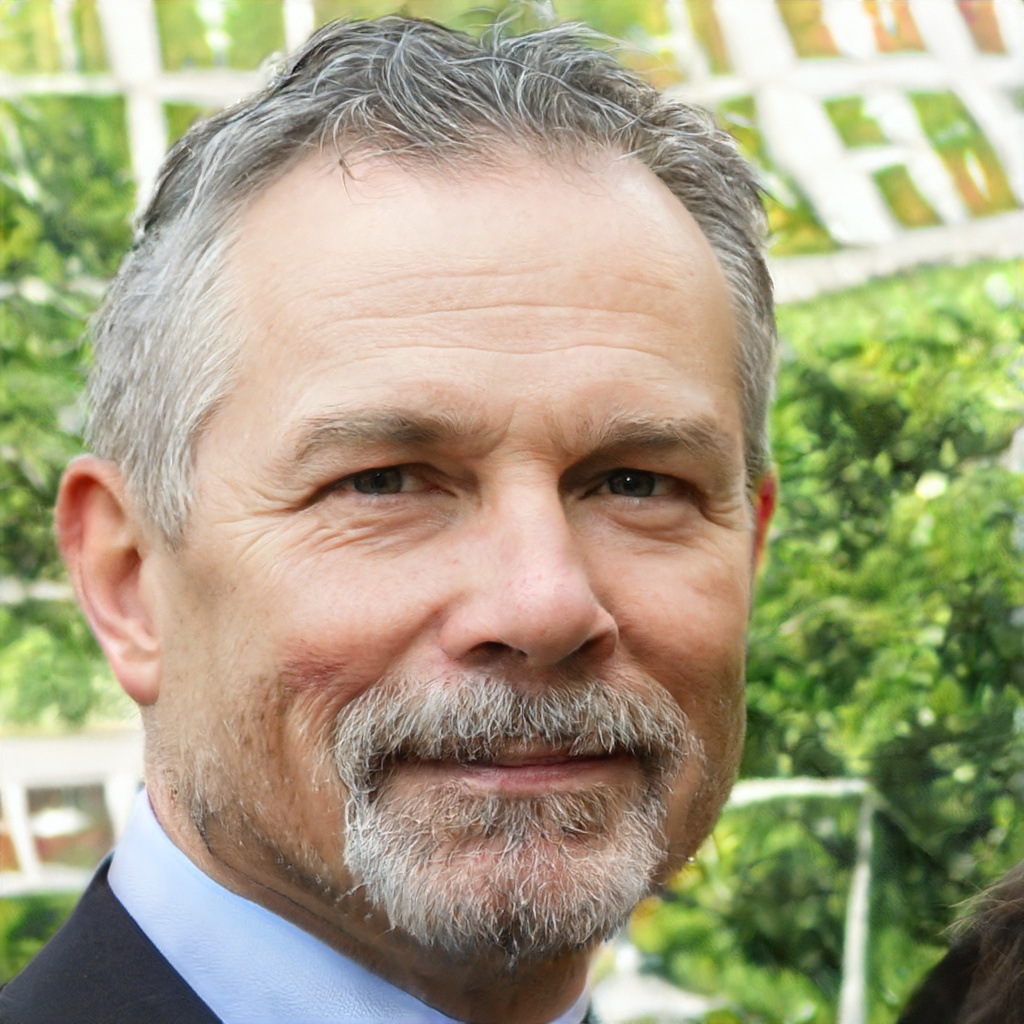}\vspace{2mm}%
\caption{
Four hand-picked examples illustrating the image quality and diversity achievable using StylegGAN2 (config \arch{f}).
}
\vspace{\hh}
\label{fig:ffhqcherrypick}
\end{figure*}
}

\newcommand{\figfrequencies}{
\begin{figure*}
\renewcommand{\h}{0.245\linewidth}
\renewcommand{\hh}{0.495\linewidth}
\includegraphics[width=\h]{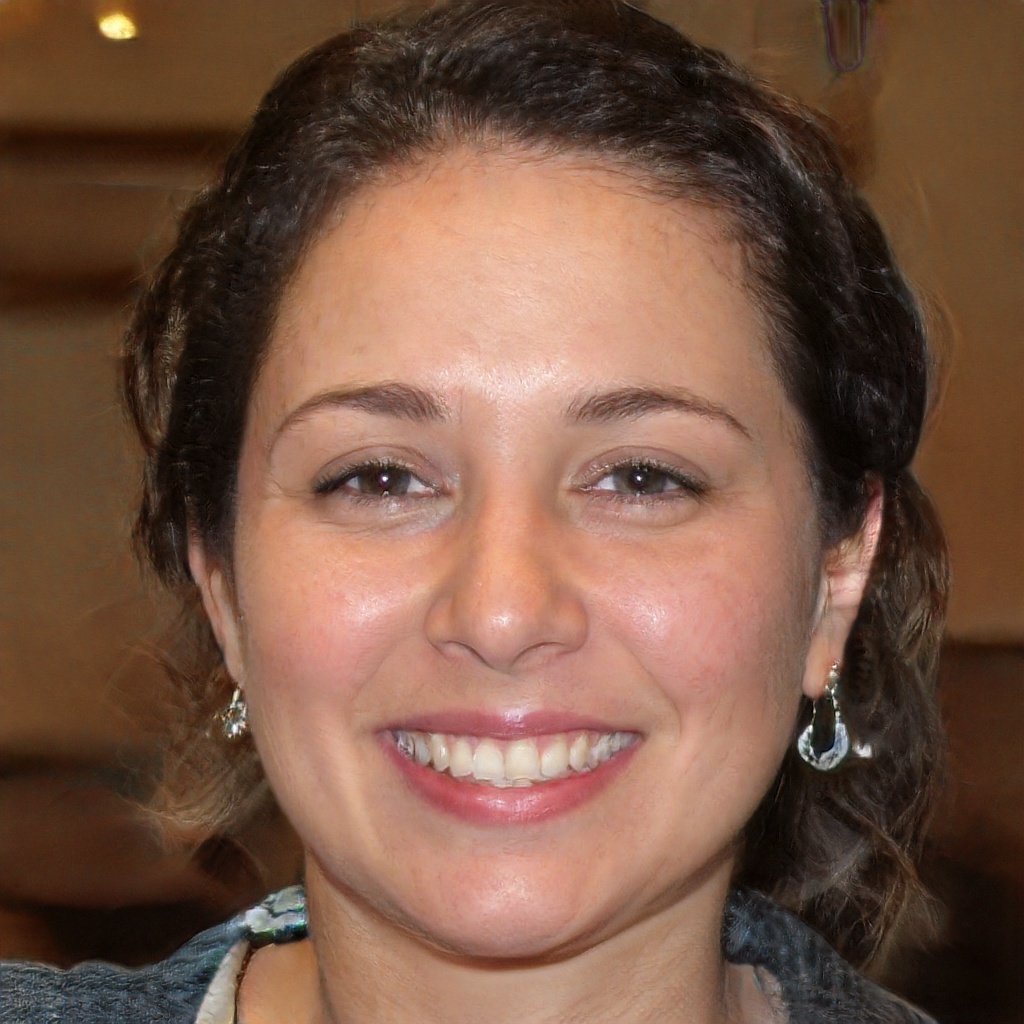}\hfill%
\includegraphics[width=\h]{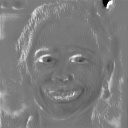}\hfill\hfill%
\includegraphics[width=\h]{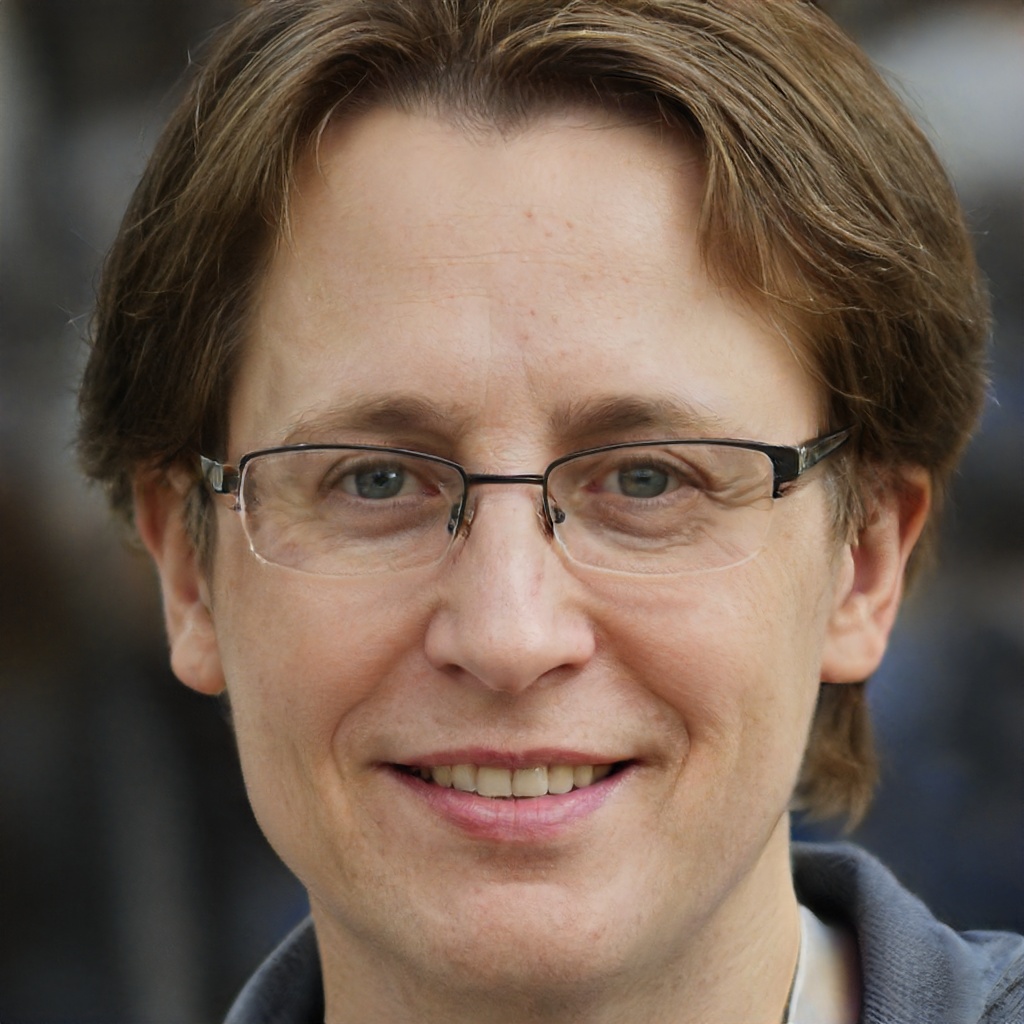}\hfill%
\includegraphics[width=\h]{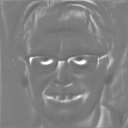}\vspace{-1mm}\\%
\footnotesize
\makebox[\h]{Generated image}\hfill%
\makebox[\h]{Feature map $128^2$}\hfill\hfill%
\makebox[\h]{Generated image}\hfill%
\makebox[\h]{Feature map $128^2$}\vspace{2mm}\\%
\small
\makebox[\hh]{(a) Progressive growing (config \arch{a})}\hfill\hfill%
\makebox[\hh]{(b) Without progressive growing (config \arch{f})}\\
\vspace*{-1mm}%
\caption{
Progressive growing leads to significantly higher frequency content in the intermediate layers. This compromises shift-invariance of the network and makes it harder to localize features precisely in the higher-resolution layers.
}
\label{fig:frequencies}
\end{figure*}
}

\newcommand{\figuncurated}{
\begin{figure*}[t]
\renewcommand{\h}{0.984\linewidth}
\setlength{\lineskip}{2mm}
\footnotesize
\rotatebox{90}{\makebox[61mm][c]{\textsc{FFHQ}}}\hfill%
\includegraphics[width=\h]{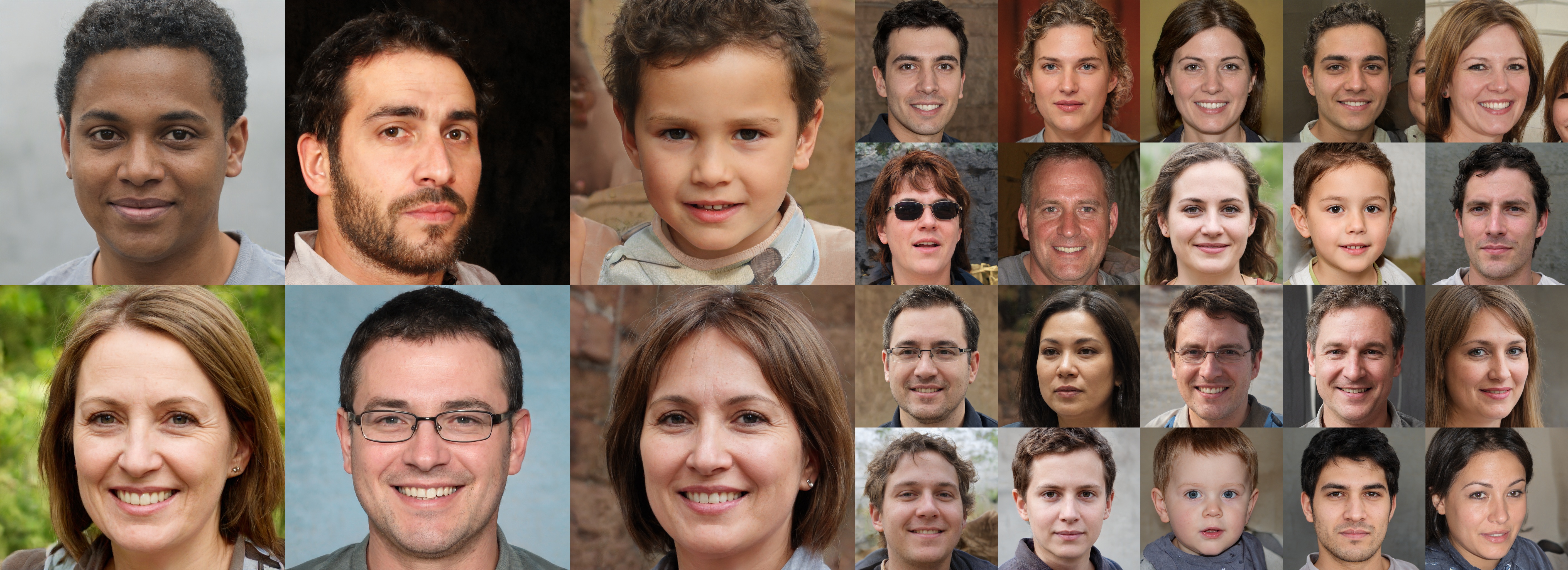}\\
\rotatebox{90}{\makebox[46mm][c]{\textsc{LSUN Car}}}\hfill%
\includegraphics[width=\h]{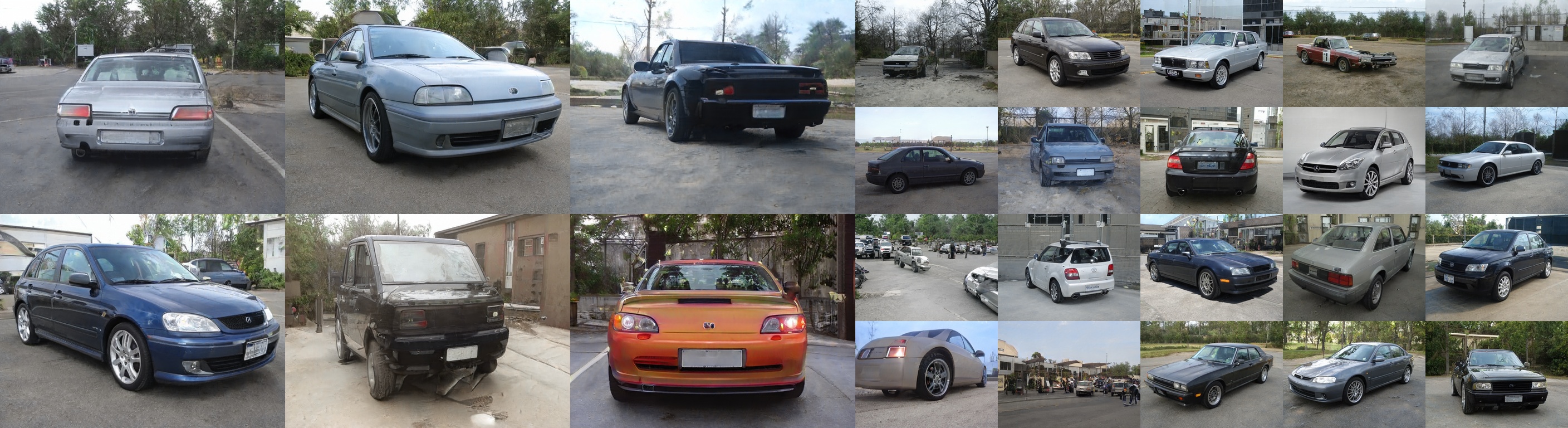}\\
\rotatebox{90}{\makebox[30mm][c]{\textsc{LSUN Cat}}}\hfill%
\includegraphics[width=\h]{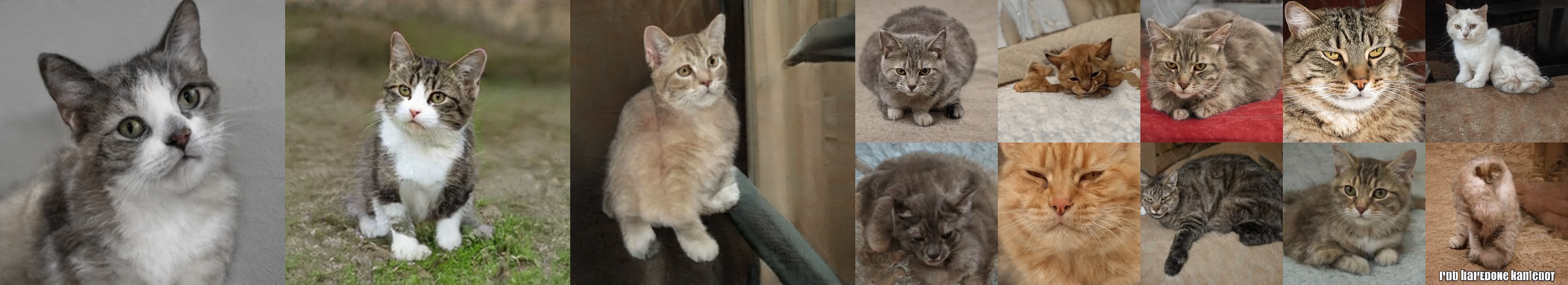}\\
\rotatebox{90}{\makebox[30mm][c]{\textsc{LSUN Church}}}\hfill%
\includegraphics[width=\h]{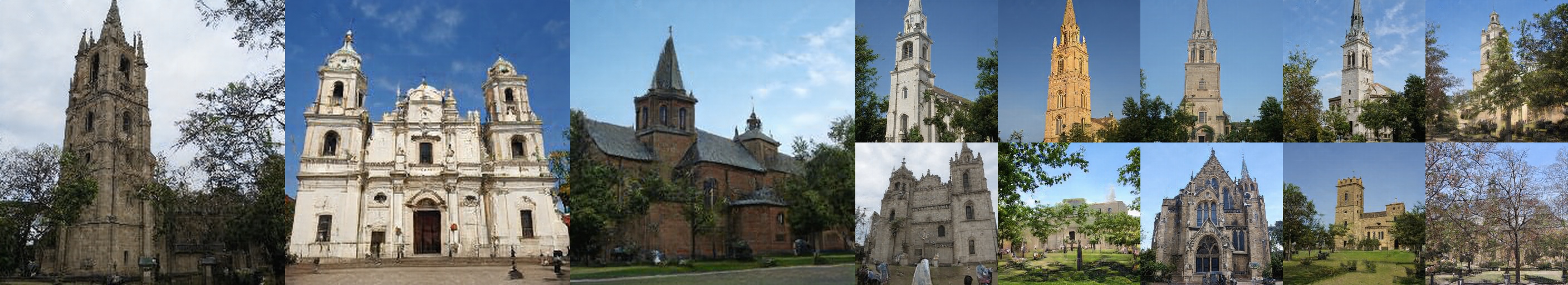}\\
\rotatebox{90}{\makebox[30mm][c]{\textsc{LSUN Horse}}}\hfill%
\includegraphics[width=\h]{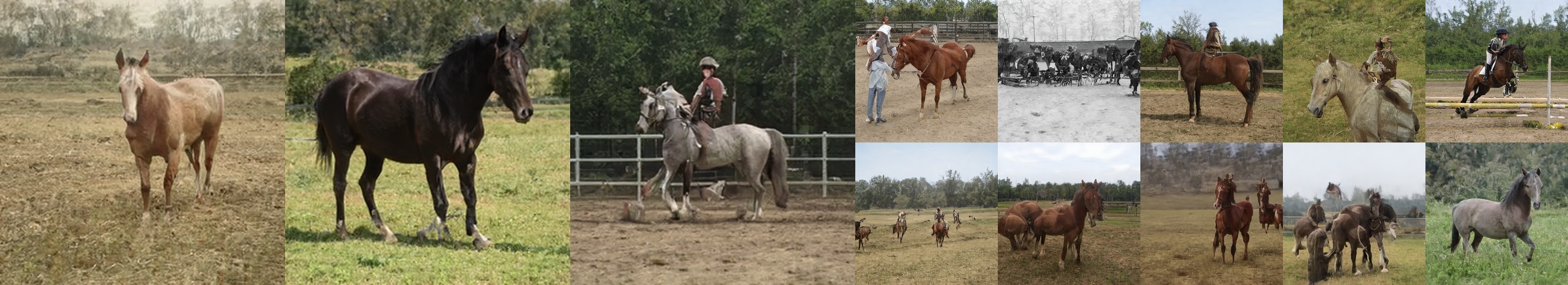}\\
\caption{%
Uncurated results for each dataset used in \reftabmains{}.
The images correspond to random outputs produced by our generator (config \arch{f}), with truncation applied at all resolutions using $\psi=0.5$ \cite{Karras2018}.
}
\label{fig:uncurated}
\end{figure*}
}

\newcommand{\figeigenvalues}{
\begin{figure}[t]
\includegraphics[width=0.49\linewidth]{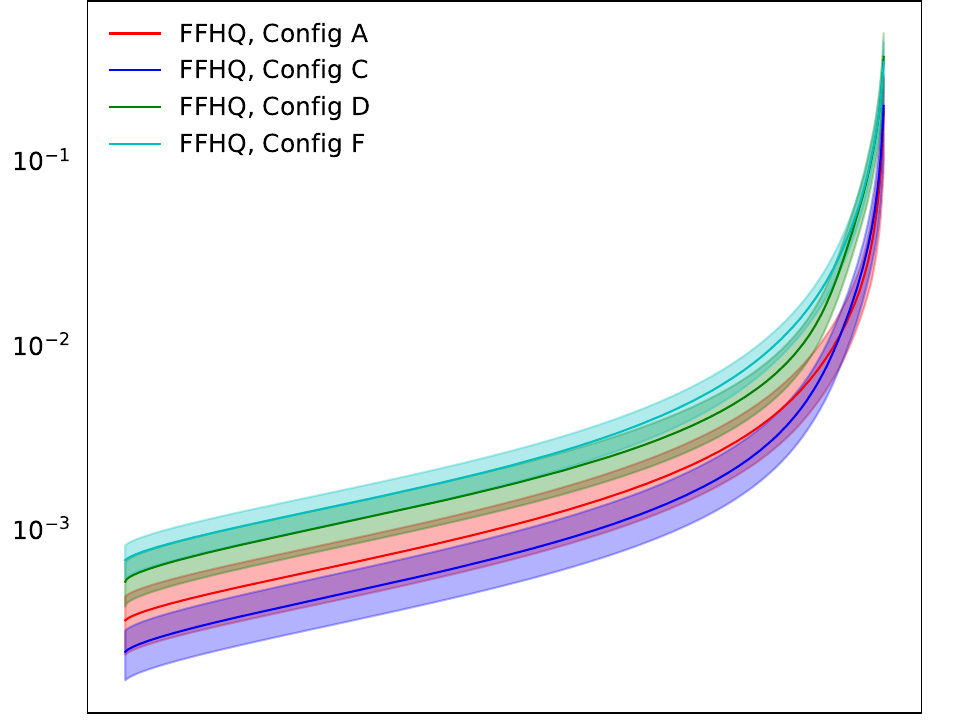}\hfill
\includegraphics[width=0.49\linewidth]{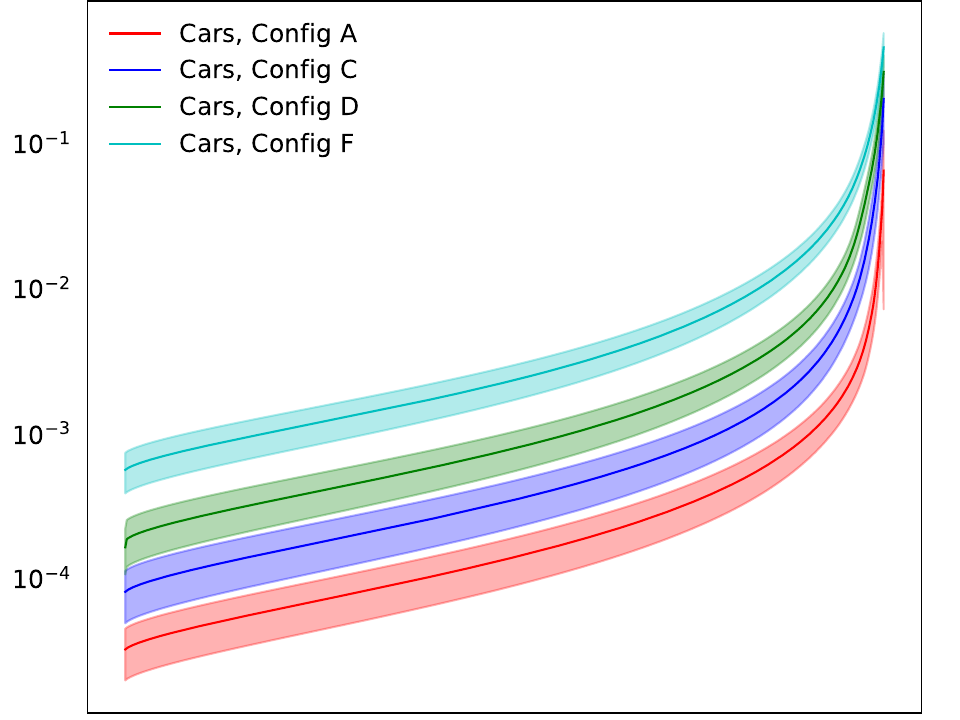}
\caption{
The mean and standard deviation of the magnitudes of sorted singular values of the Jacobian matrix evaluated at random latent space points $\ww$, with largest eigenvalue normalized to $1$. In both datasets, path length regularization (Config D) and novel architecture (Config F) exhibit better conditioning; notably, the effect is more pronounced in the Cars dataset that contains much more variability, and where path length regularization has a relatively stronger effect on the PPL metric (\reftabmain{}).
}
\label{fig:eigenvalues}
\end{figure}
}

\newcommand{\fignoiseregcars}{
\begin{figure}[t]
\renewcommand{\h}{0.2475\linewidth}
\renewcommand{\hh}{0.498333\linewidth}
\footnotesize
\makebox[\hh]{Generated target image}\hfill%
\makebox[\hh]{Real target image}\\%
\includegraphics[width=\hh]{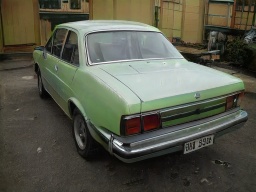}\hfill%
\includegraphics[width=\hh]{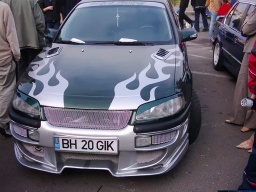}\\%
\includegraphics[width=\h]{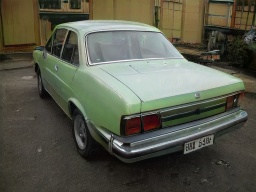}\hfill%
\includegraphics[width=\h]{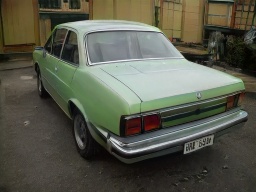}\hfill%
\includegraphics[width=\h]{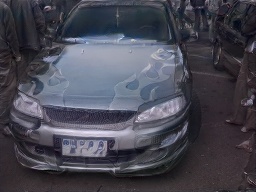}\hfill%
\includegraphics[width=\h]{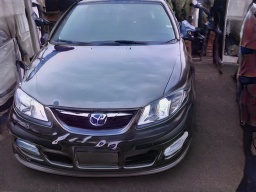}\\
\includegraphics[width=\h]{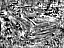}\hfill%
\includegraphics[width=\h]{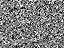}\hfill%
\includegraphics[width=\h]{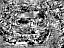}\hfill%
\includegraphics[width=\h]{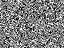}\\
\includegraphics[width=\h]{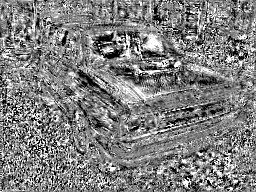}\hfill%
\includegraphics[width=\h]{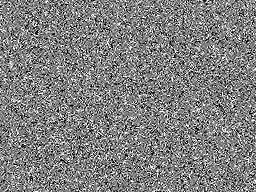}\hfill%
\includegraphics[width=\h]{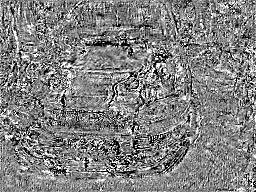}\hfill%
\includegraphics[width=\h]{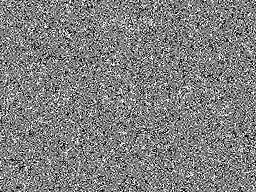}\\
\footnotesize
\makebox[\h]{No noise}\hfill%
\makebox[\h]{With noise}\hfill%
\makebox[\h]{No noise}\hfill%
\makebox[\h]{With noise}\\%
\makebox[\h]{regularization}\hfill%
\makebox[\h]{regularization}\hfill%
\makebox[\h]{regularization}\hfill%
\makebox[\h]{regularization}\\%
\caption{%
Effect of noise regularization in latent-space projection where we also optimize the contents of the noise inputs of the synthesis network.
Top to bottom: target image, re-synthesized image, contents of two noise maps at different resolutions.
When regularization is turned off in this test, we only normalize the noise maps to zero mean and unit variance, which leads the optimization to sneak signal into the noise maps.
Enabling the noise regularization prevents this. The model used here corresponds to configuration~\arch{f} in \reftabmain{}.
}
\label{fig:noiseregcars}
\end{figure}
}

\newcommand{\tabspectralnorm}{
\begin{table}[t]
\FINAL{%
\centering%
\newcolumntype{x}{>{\centering\arraybackslash\hspace{0pt}}p{4.4mm}}%
\newcolumntype{y}{>{\centering\arraybackslash\hspace{0pt}}p{6.4mm}}%
\newcommand{\YES}{\checkmark}%
\newcommand{\NO}{--}%
\footnotesize{%
\begin{tabular}{|l|xxxx|yyyy|}
\hline
&
\makebox[0mm][c]{SN-G}              &   %
\makebox[0mm][c]{SN-D}              &   %
\makebox[0mm][c]{Demod}             &   %
\makebox[0mm][c]{P.reg}             &   %
\makebox[0mm][c]{FID $\downarrow$}  &   %
\makebox[0mm][c]{PPL $\downarrow$}  &   %
\makebox[0mm][c]{Pre. $\uparrow$}   &   %
\makebox[0mm][c]{Rec. $\uparrow$}   \\  %
\hline
1 & \NO  & \NO  & \YES & \YES   & \textbf{2.83} & 145.0          & 0.689          & \textbf{0.492} \\   %
\hline
2 & \NO  & \YES & \YES & \YES   & 2.98          & 131.4          & 0.700          & 0.469          \\   %
3 & \YES & \YES & \YES & \YES   & 3.40          & \textbf{130.9} & \textbf{0.720} & 0.435          \\   %
4 & \YES & \YES & \NO  & \YES   & 3.38          & 162.6          & 0.705          & 0.468          \\   %
5 & \YES & \YES & \NO  & \NO    & 3.33          & 394.9          & 0.705          & 0.463          \\   %
6 & \YES & \NO  & \NO  & \YES   & 3.36          & 217.1          & 0.695          & 0.464          \\   %
7 & \YES & \NO  & \NO  & \NO    & 3.22          & 394.4          & 0.692          & 0.489          \\   %
\hline
\end{tabular}\vspace{2mm}}
\caption{
Effect of spectral normalization with \textsc{FFHQ} at $1024^2$.
The first row corresponds to StyleGAN2, i.e., config \arch{f} in \reftabmain{}.
In the subsequent rows, we enable spectral normalization in the generator (\emph{SN-G}) and in the discriminator (\emph{SN-D}).
We also test the training without weight demodulation (\emph{Demod}) and path length regularization (\emph{P.reg}).
All of these configurations are highly detrimental to FID, as well as to Recall.
$\uparrow$ indicates that higher is better, and $\downarrow$ that lower is better.
}\vspace{-2mm}
\label{tab:spectralnorm}
}
\end{table}
}

\newcommand{\tabpower}{
\begin{table}[t]
\FINAL{%
\centering%
\footnotesize%
\newcolumntype{x}{>{\centering\arraybackslash\hspace{0pt}}p{23mm}}%
\begin{tabular}{|l@{\hspace{2.9mm}}|x|x|}
\hline
\textbf{Item}           & \textbf{GPU years (Volta)} & \textbf{Electricity (MWh)}\\\hline
Initial exploration     & 20.25     & 58.94     \\
Paper exploration       & 13.71     & 31.49     \\
FFHQ config~\arch{f}    & \s0.23    & \s0.68    \\
Other runs in paper     & \s7.20    & 16.77     \\
Backup runs left out    & \s4.73    & 12.08     \\
Video, figures, etc.    & \s0.31    & \s0.82    \\
Public release          & \s4.62    & 10.82     \\\hline
Total                   & 51.05     & 131.61    \\\hline
\end{tabular}\vspace{3mm}
\caption{
Computational effort expenditure and electricity consumption data for this project. The unit for computation is GPU-years on a single NVIDIA V100 GPU\,---\,it would have taken approximately 51 years to execute this project using a single GPU. See the text for additional details about the computation and energy consumption estimates.
 \emph{Initial exploration} includes all training runs after the release of StyleGAN~\cite{Karras2018} that affected our decision to start this project.
 \emph{Paper exploration} includes all training runs that were done specifically for this project, but were not intended to be used in the paper as-is.
 \emph{FFHQ config F} refers to the training of the final network. This is approximately the cost of training the network for another dataset without hyperparameter tuning.
 \emph{Other runs in paper} covers the training of all other networks shown in the paper.
 \emph{Backup runs left out} includes the training of various networks that could potentially have been shown in the paper, but were ultimately left out to keep the exposition more focused.
 \emph{Video, figures, etc.} includes computation that was spent on producing the images and graphs in the paper, as well as on the result video.
 \emph{Public release} covers testing, benchmarking, and large-scale image dumps related to the public release.
}
\label{tab:power}
}
\end{table}
}

\ifcvprfinal\fi
\begin{document}

\title{Analyzing and Improving the Image Quality of StyleGAN}

\author{Tero Karras\\
NVIDIA%
\and
Samuli Laine\\
NVIDIA%
\and
Miika Aittala\\
NVIDIA%
\and
Janne Hellsten\\
NVIDIA%
\and
Jaakko Lehtinen\\
NVIDIA and Aalto University%
\and
Timo Aila\\
NVIDIA%
}

\maketitle

\ifcvprfinal\thispagestyle{empty}\fi
\ifarxiv
\pagestyle{plain}\thispagestyle{plain}
\fi

\ifforceplain\thispagestyle{plain}\pagestyle{plain}\fi

\ifarxiv
	\newcommand{\refappimagequality}{Appendix~\ref{app:imagequality}}
	\newcommand{\refappimplementation}{Appendix~\ref{app:implementation}}
	\newcommand{\refappregularization}{Appendix~\ref{app:regularization}}
	\newcommand{\refappprojectiondetails}{Appendix~\ref{app:projectiondetails}}
	\newcommand{\refapplsuncomparison}{Figures~\ref{fig:catscomparison} and~\ref{fig:carscomparison}}
	\newcommand{\refappspectralnorm}{Appendix~\ref{app:spectralnorm}}
	\newcommand{\refapppower}{Appendix~\ref{app:power}}
\else
	\newcommand{\refappimagequality}{Appendix~A}
	\newcommand{\refappimplementation}{Appendix~B}
	\newcommand{\refappregularization}{Appendix~C}
	\newcommand{\refappprojectiondetails}{Appendix~D}
	\newcommand{\refapplsuncomparison}{Figures~3 and 4 in the Supplement}
	\newcommand{\refappspectralnorm}{Appendix~E}
	\newcommand{\refapppower}{Appendix~F}
\fi
\newcommand{\reftablsun}{Table~\ref{tab:lsun}}

\begin{abstract}

The style-based GAN architecture (StyleGAN) yields state-of-the-art results in data-driven unconditional generative image modeling. 
We expose and analyze several of its characteristic artifacts, and propose changes in both model architecture and training methods to address them.
In particular, we redesign the generator normalization, revisit progressive growing, and regularize the generator to encourage good conditioning in the mapping from latent codes to images. 
In addition to improving image quality, this path length regularizer yields the additional benefit that the generator becomes significantly easier to invert.
\FINAL{This makes it possible to reliably attribute a generated image to a particular network.} 
We furthermore visualize how well the generator utilizes its output resolution, and identify a capacity problem, motivating us to train larger models for additional quality improvements. 
Overall, our improved model redefines the state of the art in unconditional image modeling, both in terms of existing distribution quality metrics as well as perceived image quality.

\end{abstract}

\section{Introduction}

The resolution and quality of images produced by generative methods, especially generative adversarial networks (GAN) \cite{Goodfellow2014}, are improving rapidly \cite{Karras2017,Miyato2018B,Brock2018}. 
The current state-of-the-art method for high-resolution image synthesis is StyleGAN \cite{Karras2018}, which has been shown to work reliably on a variety of datasets. 
Our work focuses on fixing its characteristic artifacts and improving the result quality further.

The distinguishing feature of StyleGAN \cite{Karras2018} is its unconventional generator architecture. Instead of feeding the input latent code $\zz \in \ZZ$ only to the beginning of a the network, the \emph{mapping network} $f$ first transforms it to an intermediate latent code $\ww \in \WW$.
Affine transforms then produce \emph{styles} that control the layers of the \emph{synthesis network} $g$ via adaptive instance normalization (AdaIN) \cite{Huang2017,Dumoulin2016,Ghiasi2017,Dumoulin2018}.
Additionally, stochastic variation is facilitated by providing additional random noise maps to the synthesis network.
It has been demonstrated \cite{Karras2018,Shen2019} %
that this design allows the intermediate latent space $\WW$ to be much less entangled than the input latent space $\ZZ$.
In this paper, we focus all analysis solely on $\WW$, as it is the relevant latent space from the synthesis network's point of view.

\figdroplets %

Many observers have noticed characteristic artifacts in images generated by StyleGAN \cite{Bergstrom2019}. We identify two causes for these artifacts, and describe changes in architecture and training methods that eliminate them.
First, we investigate the origin of common blob-like artifacts, and find that the generator creates them to circumvent a design flaw in its architecture. 
In Section~\ref{sec:in}, we redesign the normalization used in the generator, which removes the artifacts.
Second, we analyze artifacts related to progressive growing \cite{Karras2017} that has been highly successful in stabilizing high-resolution GAN training.
We propose an alternative design that achieves the same goal\,---\,training starts by focusing on
low-resolution images and then progressively shifts focus to higher and higher resolutions\,---\,without changing the network topology during training.
This new design also allows us to reason about the effective resolution of the generated images, which turns out to be lower than expected, motivating a capacity increase (Section~\ref{sec:growing}).

Quantitative analysis of the quality of images produced using generative methods continues to be a challenging topic. 
Fr\'echet inception distance (FID) \cite{Heusel2017} measures differences in the density of two distributions in the high-dimensional feature space of an InceptionV3 classifier \cite{simonyan2014}. 
Precision and Recall (P\&R) \cite{Sajjadi2018,Tuomas2019} provide additional visibility by explicitly quantifying the percentage of generated images that are similar to training data and the percentage of training data that can be generated, respectively. 
We use these metrics to quantify the improvements.

Both FID and P\&R are based on classifier networks that have recently been shown to focus on textures rather than shapes \cite{Geirhos2018}, and consequently, the metrics do not accurately capture all aspects of image quality.
We observe that the perceptual path length (PPL) metric \cite{Karras2018}, originally introduced as a method for estimating the quality of latent space interpolations, correlates with consistency and stability of shapes.
Based on this, we regularize the synthesis network to favor smooth mappings (Section~\ref{sec:Greg}) and achieve a clear improvement in quality.
To counter its computational expense, we also propose executing all regularizations less frequently, observing that this can be done without compromising effectiveness.

Finally, we find that projection of images to the latent space $\WW$ works significantly better with the new, path-length regularized \FINAL{StyleGAN2 generator than with the original StyleGAN. 
This makes it easier to attribute a generated image to its source (Section~\ref{sec:proj}).}

Our implementation and trained models are available at \texttt{\small https://github.com/NVlabs/stylegan2}

\section{Removing normalization artifacts}
\label{sec:in}

\figarch %

We begin by observing that most images generated by StyleGAN exhibit characteristic blob-shaped artifacts that resemble water droplets.
As shown in Figure~\ref{fig:droplets}, even when the droplet may not be obvious in the final image, it is present in the intermediate feature maps of the generator.%
\footnote{In rare cases (perhaps 0.1\% of images) the droplet is missing, leading to severely corrupted images. See \refappimagequality{} for details.}
The anomaly starts to appear around 64$\times$64 resolution, is present in all feature maps, and becomes progressively stronger at higher resolutions.
The existence of such a consistent artifact is puzzling, as the discriminator should be able to detect it.

We pinpoint the problem to the AdaIN operation that normalizes the mean and variance of each feature map separately, thereby potentially destroying any information found in the magnitudes of the features relative to each other.
We hypothesize that the droplet artifact is a result of the generator intentionally sneaking signal strength information past instance normalization:
by creating a strong, localized spike that dominates the statistics, the generator can effectively scale the signal as it likes elsewhere.
Our hypothesis is supported by the finding that when the normalization step is removed from the generator, as detailed below, the droplet artifacts disappear completely.

\subsection{Generator architecture revisited}

We will first revise several details of the StyleGAN generator to better facilitate our redesigned normalization. These changes have either a neutral or small positive effect on their own in terms of quality metrics.

Figure~\ref{fig:arch}a shows the original StyleGAN synthesis network $g$ \cite{Karras2018}, and in Figure~\ref{fig:arch}b we expand the diagram to full detail by showing the weights and biases and breaking the AdaIN operation to its two constituent parts: normalization and modulation. This allows us to re-draw the conceptual gray boxes so that each box indicates the part of the network where one style is active (i.e., ``style block'').
Interestingly, the original StyleGAN applies bias and noise within the style block, causing their relative impact to be inversely proportional to the current style's magnitudes. We observe that more predictable results are obtained by moving these operations outside the style block, where they operate on normalized data. Furthermore, we notice that after this change it is sufficient for the normalization and modulation to operate on the standard deviation alone (i.e., the mean is not needed).
The application of bias, noise, and normalization to the constant input can also be safely removed without observable drawbacks. 
This variant is shown in Figure~\ref{fig:arch}c, and serves as a starting point for our redesigned normalization.

\subsection{Instance normalization revisited}

\FINAL{One of the main strengths of StyleGAN is the ability to control the generated images via \emph{style mixing}, i.e., by feeding a different latent $\ww$ to different layers at inference time. In practice, style modulation may amplify certain feature maps by an order of magnitude or more.
For style mixing to work, we must explicitly counteract this amplification on a per-sample basis\,---\,otherwise the subsequent layers would not be able to operate on the data in a meaningful way.

If we were willing to sacrifice scale-specific controls (see video), we could simply remove the normalization, thus removing the artifacts and also improving FID slightly \cite{Tuomas2019}.
We will now propose a better alternative that removes the artifacts while retaining full controllability.}
The main idea is to base normalization on the \emph{expected} statistics of the incoming feature maps, but without explicit forcing.

Recall that a style block in Figure~\ref{fig:arch}c consists of modulation, convolution, and normalization. 
Let us start by considering the effect of a modulation followed by a convolution. The modulation scales each input feature map of the convolution based on the incoming style, which can alternatively be implemented by scaling the convolution weights:
\begin{equation}
\label{eq:modulation}
w'_{ijk} = s_i \cdot w_{ijk},
\end{equation}
where $w$ and $w'$ are the original and modulated weights, respectively, $s_i$ is the scale corresponding to the $i$th input feature map, and $j$ and $k$ enumerate the output feature maps and spatial footprint of the convolution, respectively.

Now, the purpose of instance normalization is to essentially remove the effect of $s$ from the statistics of the convolution's output feature maps.
We observe that this goal can be achieved more directly. %
Let us assume that the input activations are i.i.d.~random variables with unit standard deviation. After modulation and convolution, the output activations have standard deviation of
\begin{equation}
\sigma_j = \sqrt{\raisebox{0mm}[4.0mm][2.5mm]{$\underset{i,k}{{}\displaystyle\sum{}}$} {w'_{ijk}}^2},
\label{eq:b}
\end{equation}
i.e.,~the outputs are scaled by the $L_2$ norm of the corresponding weights. The subsequent normalization aims to restore the outputs back to unit standard deviation.
Based on Equation~\ref{eq:b}, this is achieved if we scale (``demodulate'') each output feature map $j$ by  $1/\sigma_j$. Alternatively, we can again bake this into the convolution weights:
\begin{equation}
w''_{ijk} = w'_{ijk} \bigg/ \sqrt{\raisebox{0mm}[4.0mm][2.5mm]{$\underset{i,k}{{}\displaystyle\sum{}}$} {w'_{ijk}}^2 + \epsilon},
\label{eq:demodulation}
\end{equation}
where $\epsilon$ is a small constant to avoid numerical issues.

\tabmain %

We have now baked the entire style block to a single convolution layer whose weights are adjusted based on $s$ using Equations \ref{eq:modulation} and \ref{eq:demodulation} (Figure~\ref{fig:arch}d). Compared to instance normalization, our demodulation technique is weaker because it is based on statistical assumptions about the signal instead of actual contents of the feature maps. Similar statistical analysis has been extensively used in modern network initializers \cite{Glorot2010,He2015}, but we are not aware of it being previously used as a replacement for data-dependent normalization.
Our demodulation is also related to weight normalization \cite{Salimans2016A} that performs the same calculation as a part of reparameterizing the weight tensor.
Prior work has identified weight normalization as beneficial in the context of GAN training \cite{Xiang2017}.

Our new design removes the characteristic artifacts (Figure~\ref{fig:VIN}) while retaining full controllability, as demonstrated in the accompanying video.
FID remains largely unaffected (Table~\ref{tab:main}, rows~\arch{a},~\arch{b}), but there is a notable shift from precision to recall.
We argue that this is generally desirable, since recall can be traded into precision via truncation, whereas the opposite is not true \cite{Tuomas2019}.
In practice our design can be implemented efficiently using grouped convolutions, as detailed in \refappimplementation{}.
To avoid having to account for the activation function in Equation~\ref{eq:demodulation}, we scale our activation functions so that they retain the expected signal variance.

\section{Image quality and generator smoothness}
\label{sec:Greg}

\figVIN  %

While GAN metrics such as FID or Precision and Recall (P\&R) successfully capture many aspects of the generator, they continue to have somewhat of a blind spot for image quality. For an example, refer to 
\refapplsuncomparison{} that contrast generators with identical FID and P\&R scores but markedly different overall quality.%
\footnote{%
We believe that the key to the apparent inconsistency lies in the particular choice of feature space rather than the foundations of FID or P\&R.
It was recently discovered that classifiers trained using ImageNet \cite{Russakovsky2014} tend to base their decisions much more on texture than shape \cite{Geirhos2018}, while humans strongly focus on shape \cite{Landau1988}. 
This is relevant in our context because FID and P\&R use high-level features from InceptionV3 \cite{simonyan2014} and VGG-16 \cite{simonyan2014}, respectively, which were trained in this way and are thus expected to be biased towards texture detection. 
As such, images with, e.g., strong cat textures may appear more similar to each other than a human observer would agree, thus partially compromising density-based metrics (FID) and manifold coverage metrics (P\&R).
}

\figpplimages %
\figpplhistograms %

We observe a correlation between perceived image quality and perceptual path length (PPL) \cite{Karras2018}, a metric that was originally introduced for quantifying the smoothness of the mapping from a latent space to the output image by measuring average LPIPS distances~\cite{Zhang2018metric} between generated images under small perturbations in latent space.
Again consulting \refapplsuncomparison{}, a smaller PPL (smoother generator mapping) appears to correlate with higher overall image quality, whereas other metrics are blind to the change.
\FINAL{Figure~\ref{fig:pplimages} examines this correlation more closely through per-image PPL scores on \textsc{LSUN Cat}, computed by sampling the latent space around $\ww \sim f(\mathbf{z})$. Low scores are indeed indicative of high-quality images, and vice versa.
}
Figure~\ref{fig:pplhistograms}a shows the corresponding histogram and reveals the long tail of the distribution.
The overall PPL for the model is simply the expected value of these per-image PPL scores. %
\FINAL{We always compute PPL for the entire image, as opposed to Karras~et~al.~\cite{Karras2018} who use a smaller central crop.}

It is not immediately obvious why a low PPL should correlate with image quality.
We hypothesize that during training, as the discriminator penalizes broken images, the most direct way for the generator to improve is to effectively stretch the region of latent space that yields good images.
This would lead to the low-quality images being squeezed into small latent space regions of rapid change. %
While this improves the average output quality in the short term, the accumulating distortions impair the training dynamics and consequently the final image quality.

\FINAL{
Clearly, we cannot simply encourage minimal PPL since that would guide the generator toward a degenerate solution with zero recall.
Instead, we will describe a new regularizer that aims for a smoother generator mapping without this drawback.}
As the resulting regularization term is somewhat expensive to compute, we first describe a general optimization that applies to \FINAL{any regularization technique.}

\subsection{Lazy regularization}
\label{sec:lazyreg}
Typically the main loss function (e.g.,~logistic loss~\cite{Goodfellow2014}) and regularization terms (e.g.,~$R_1$~\cite{Mescheder2018}) are written as a single expression and are thus optimized simultaneously.
We observe that the regularization terms can be computed less frequently than the main loss function, thus greatly diminishing their computational cost and the overall memory usage. 
Table~\ref{tab:main}, row~\arch{c} shows that no harm is caused when $R_1$ regularization is performed only once every 16 minibatches, and we adopt the same strategy for our new regularizer as well.
\refappimplementation{} gives implementation details.

\subsection{Path length regularization}
\label{sec:pathreg}

\FINAL{We would like to encourage that a fixed-size step in $\WW$ results in a non-zero, fixed-magnitude change in the image. 
We can measure the deviation from this ideal empirically by stepping into random directions in the image space and observing the corresponding $\ww$ gradients.
These gradients should have close to an equal length regardless of $\ww$ or the image-space direction, indicating that the mapping from the latent space to image space is well-conditioned \cite{Odena2018}.
}

At a single $\ww \in \WW$, the local metric scaling properties of the generator mapping $g(\ww) : \WW \mapsto \YY$  are captured by the Jacobian matrix $\mathbf{J}_\ww = {\partial g(\ww)}/{\partial \ww}$. %
Motivated by the desire to preserve the expected lengths of vectors regardless of the direction, we formulate our regularizer as
\begin{equation}
\expectation_{\ww, \yy \sim \mathcal{N}(0, \mathbf{I})} \left(\left\lVert \mathbf{J}_\ww^T \yy\right\rVert_2 - a\right)^2,
\end{equation}
where $\yy$ are random images with normally distributed pixel intensities, and $\ww \sim f(\mathbf{z})$, where $\mathbf{z}$ are normally distributed. We show in \refappregularization{} that, in high dimensions, this prior is minimized when $\mathbf{J}_\ww$ is orthogonal (up to a global scale) at any $\ww$. An orthogonal matrix preserves lengths and introduces no squeezing along any dimension.

To avoid explicit computation of the Jacobian matrix, we use the identity $\JJT_\ww \yy = \nabla_\ww (g(\ww)\cdot \yy)$, which is efficiently computable using standard backpropagation \cite{Dauphin2015}. The constant $a$ is set dynamically during optimization as the long-running exponential moving average of the lengths $\lVert\JJT_\ww \yy\rVert_2$, allowing the optimization to find a suitable global scale by itself.

Our regularizer is closely related to the Jacobian clamping regularizer presented by Odena et~al.~\cite{Odena2018}. Practical differences include that we compute the products $\JJT_\ww \yy$ analytically whereas they use finite differences for estimating $\mathbf{J}_\ww \boldsymbol{\delta}$ with $\ZZ \ni \boldsymbol{\delta} \sim \mathcal{N}(0, \mathbf{I})$.
It should be noted that spectral normalization \cite{Miyato2018B} of the generator \cite{Zhang2018sagan} only constrains the largest singular value, posing no constraints on the others and hence not necessarily leading to better conditioning.
\FINAL{We find that enabling spectral normalization in addition to our contributions\,---\,or instead of them\,---\,invariably compromises FID, as detailed in \refappspectralnorm{}.}

In practice, we notice that path length regularization leads to more reliable and consistently behaving models, making architecture exploration easier.
\FINAL{We also observe that the smoother generator is significantly easier to invert (Section~\ref{sec:proj}).}
Figure~\ref{fig:pplhistograms}b shows that path length regularization clearly tightens the distribution of per-image PPL \FINAL{scores, without pushing the mode to zero.}
\FINAL{However, Table~\ref{tab:main}, row~\arch{d} points toward a tradeoff between FID and PPL in datasets that are less structured than \textsc{FFHQ}.}

\section{Progressive growing revisited}
\label{sec:growing}

\figpgartifacts %

Progressive growing \cite{Karras2017} has been very successful in stabilizing high-resolution image synthesis, but it causes its own characteristic artifacts. 
The key issue is that the progressively grown generator appears to have a strong location preference for details; 
the accompanying video shows that when features like teeth or eyes should move smoothly over the image, they may instead remain stuck in place before jumping to the next preferred location. 
Figure~\ref{fig:pgartifacts} shows a related artifact. 
We believe the problem is that in progressive growing each resolution serves momentarily as the output resolution, forcing it to generate maximal frequency details, which then leads to the trained network to have excessively high frequencies in the intermediate layers, compromising shift invariance \cite{zhang2019}. \refappimagequality{} shows an example.
These issues prompt us to search for an alternative formulation that would retain the benefits of progressive growing without the drawbacks.

\subsection{Alternative network architectures}
\label{sec:altarch}

\figGDarch %

While StyleGAN uses simple feedforward designs in the generator (synthesis network) and discriminator, there is a vast body of work dedicated to the study of better network architectures.
Skip connections \cite{Ronneberger2015,Karnewar2019}, residual networks \cite{He2015b,Gulrajani2017,Miyato2018B}, and hierarchical methods \cite{Denton2015,Zhang2016,Zhang2017} have proven highly successful also in the context of generative methods.
As such, we decided to re-evaluate the network design of StyleGAN and search for an architecture that produces high-quality images without progressive growing.

Figure~\ref{fig:GDarch}a shows \FINAL{MSG-GAN \cite{Karnewar2019}}, which connects the matching resolutions of the generator and discriminator using multiple skip connections. 
The MSG-GAN generator is modified to output a mipmap \cite{Williams1983} instead of an image, and a similar representation is computed for each real image as well.
In Figure~\ref{fig:GDarch}b we simplify this design by upsampling and summing the contributions of RGB outputs corresponding to different resolutions. 
In the discriminator, we similarly provide the downsampled image to each resolution block of the discriminator.
We use bilinear filtering in all up and downsampling operations. 
In Figure~\ref{fig:GDarch}c we further modify the design to use residual connections.%
\footnote{In residual network architectures, the addition of two paths leads to a doubling of signal variance, which we cancel by multiplying with $1/\sqrt 2$. 
This is crucial for our networks, whereas in classification resnets~\cite{He2015b} the issue is typically hidden by batch normalization.}
This design is similar to LAPGAN \cite{Denton2015} without the per-resolution discriminators employed by Denton~et~al.

\tabskips %

Table~\ref{tab:skips} compares three generator and three discriminator architectures: original feedforward networks as used in StyleGAN, skip connections, and residual networks, all trained without progressive growing.
FID and PPL are provided for each of the 9 combinations.
We can see two broad trends: skip connections in the generator drastically improve PPL in all configurations, and a residual discriminator network is clearly beneficial for FID.
The latter is perhaps not surprising since the structure of discriminator resembles classifiers where residual architectures are known to be helpful. 
However, a residual architecture was harmful in the generator\,---\,the lone exception was FID in \textsc{LSUN Car} when both networks were residual. 

For the rest of the paper we use a skip generator and a residual discriminator, without progressive growing.
This corresponds to configuration~\arch{e} in Table~\ref{tab:main}, and it significantly improves FID and PPL. 

\subsection{Resolution usage}
\label{sec:resusage}

\figresolutionusage %

The key aspect of progressive growing, which we would like to preserve, is that the generator will initially focus on low-resolution features and then slowly shift its attention to finer details.
The architectures in Figure~\ref{fig:GDarch} make it possible for the generator to first output low resolution images that are not affected by the higher-resolution layers in a significant way, and later shift the focus to the higher-resolution layers as the training proceeds. Since this is not enforced in any way, the generator will do it only if it is beneficial. To analyze the behavior in practice, we need to quantify how strongly the generator relies on particular resolutions over the course of training.

Since the skip generator (Figure~\ref{fig:GDarch}b) forms the image by explicitly summing RGB values from multiple resolutions, we can estimate the relative importance of the corresponding layers by measuring how much they contribute to the final image. In Figure~\ref{fig:resolutionusage}a, we plot the standard deviation of the pixel values produced by each tRGB layer as a function of training time. We calculate the standard deviations over 1024 random samples of $\ww$ and normalize the values so that they sum to 100\%.

At the start of training, we can see that the new skip generator behaves similar to progressive growing\,---\,now achieved without changing the network topology.
It would thus be reasonable to expect the highest resolution to dominate towards the end of the training. The plot, however, shows that this fails to happen in practice, which indicates that the generator may not be able to ``fully utilize'' the target resolution. To verify this, we inspected the generated images manually and noticed that they generally lack some of the pixel-level detail that is present in the training data\,---\,the images could be described as being sharpened versions of $512^2$ images instead of true $1024^2$ images.

This leads us to hypothesize that there is a capacity problem in our networks, which we test by doubling the number of feature maps in the highest-resolution layers of both networks.%
\footnote{We double the number of feature maps in resolutions $64^2$--$1024^2$ while keeping other parts of the networks unchanged. This increases the total number of trainable parameters in the generator by 22\% (25M $\rightarrow$ 30M) and in the discriminator by 21\% (24M $\rightarrow$ 29M).}
This brings the behavior more in line with expectations: Figure~\ref{fig:resolutionusage}b shows a significant increase in the contribution of the highest-resolution layers, and Table~\ref{tab:main}, row~\arch{f} shows that FID and Recall improve markedly. 
\FINAL{The last row shows that baseline StyleGAN also benefits from additional capacity, but its quality remains far below StyleGAN2.}

\tablsun            %

Table~\ref{tab:lsun} compares StyleGAN and StyleGAN2 in four LSUN categories, again showing clear improvements in FID and significant advances in PPL.
It is possible that further increases in the size could provide additional benefits.%

\section{Projection of images to latent space}
\label{sec:proj}

Inverting the synthesis network $g$ is an interesting problem that has many applications. 
Manipulating a given image in the latent feature space requires finding a matching latent code $\ww$ for it first.
Previous research \cite{Abdal2019,Gabbay2019} suggests that instead of finding a common latent code $\ww$, the results improve if a separate $\ww$ is chosen for each layer of the generator.
The same approach was used in an early encoder implementation~\cite{puzer}. 
While extending the latent space in this fashion finds a closer match to a given image, it also enables projecting arbitrary images that should have no latent representation.
Instead, we concentrate on finding latent codes in the original, unextended latent space, as these correspond to images that the generator could have produced.

\figprojimgcarhorizontal  %
\figprojhist %

Our projection method differs from previous methods in two ways. First, we add ramped-down noise to the latent code during optimization in order to explore the latent space more comprehensively.
Second, we also optimize the stochastic noise inputs of the StyleGAN generator, regularizing them to ensure they do not end up carrying coherent signal.
The regularization is based on enforcing the autocorrelation coefficients of the noise maps to match those of unit Gaussian noise over multiple scales.
Details of our projection method can be found in \refappprojectiondetails{}.

\subsection{Attribution of generated images}

\FINAL{
Detection of manipulated or generated images is a very important task. 
At present, classifier-based methods can quite reliably detect generated images, regardless of their exact origin ~\cite{Li2018,Yu2018,Wang2019,Zhang2019ganartifacts,Wang2019b}.
However, given the rapid pace of progress in generative methods, this may not be a lasting situation.
Besides general detection of fake images, we may also consider a more limited form of the problem: being able to attribute a fake image to its specific source~\cite{Albright2019}. With StyleGAN, this amounts to checking if there exists a $\ww \in \WW$ that re-synthesis the image in question.}

We measure how well the projection succeeds by computing the LPIPS~\cite{Zhang2018metric} distance
between original and re-synthesized image as \mbox{$D_\mathrm{LPIPS}[\boldsymbol{x}, g(\tilde{g}^{-1}(\boldsymbol{x}))]$}, where $\boldsymbol{x}$ is the image being analyzed
and \mbox{$\tilde{g}^{-1}$} denotes the approximate projection operation.
Figure~\ref{fig:projhist} shows histograms of these distances for \textsc{LSUN Car} and \textsc{FFHQ} datasets using the original StyleGAN and StyleGAN2, and
Figure~\ref{fig:projimgcar} shows example projections. 
\FINAL{The images generated using StyleGAN2 can be projected into $\WW$ so well that they can be almost unambiguously attributed to the generating network.
However, with the original StyleGAN, even though it should technically be possible to find a matching latent code, it appears that the mapping from $\WW$ to images is too complex for this to succeed reliably in practice.
We find it encouraging that StyleGAN2 makes source attribution easier even though the image quality has improved significantly.}

\section{Conclusions and future work}

We have identified and fixed several image quality issues in StyleGAN, improving the quality further and considerably advancing the state of the art in several datasets.
In some cases the improvements are more clearly seen in motion, as demonstrated in the accompanying video.
\refappimagequality{} includes further examples of results obtainable using our method.
\FINAL{Despite the improved quality, StyleGAN2 makes it easier to attribute a generated image to its source.}

Training performance has also improved.
At $1024^2$ resolution, the original StyleGAN (config~\arch{a} in Table~\ref{tab:main}) trains at 37 images per second on NVIDIA DGX-1 with 8 Tesla V100 GPUs, while our config~\arch{e} trains 40\% faster at 61 img/s.
Most of the speedup comes from simplified dataflow due to weight demodulation, lazy regularization, and code optimizations.
StyleGAN2 (config~\arch{f}, larger networks) trains at 31 img/s, and is thus only slightly more expensive to train than original StyleGAN.
Its total training time was 9 days for \textsc{FFHQ} and 13 days for \textsc{LSUN Car}.

\FINAL{The entire project, including all exploration, consumed 132 MWh of electricity, of which 0.68 MWh went into training the final \textsc{FFHQ} model. In total, we used about 51 single-GPU years of computation (Volta class GPU). A more detailed discussion is available in \refapppower{}.}

In \FINAL{the} future, it could be fruitful to study further improvements to the path length regularization, e.g.,~by replacing the pixel-space $L_2$ distance with a data-driven feature-space metric.
\FINAL{Considering the practical deployment of GANs, we feel that it will be important to find new ways to reduce the training data requirements. This is especially crucial in applications where it is infeasible to acquire tens of thousands of training samples, and with datasets that include a lot of intrinsic variation.}

\vspace{-2mm}
\paragraph*{\FINAL{Acknowledgements}}
\FINAL{
We thank Ming-Yu Liu for an early review, Timo Viitanen for help with the public release, David Luebke for in-depth discussions and helpful comments, and Tero Kuosmanen for technical support with the compute infrastructure. 
}

{\small
\bibliographystyle{ieee_fullname}
\bibliography{paper}
}

\ifarxiv
\clearpage
\appendix
\ifarxiv
	\newcommand{\refsecin}{Section~\ref{sec:in}}
	\newcommand{\reffigarch}{Figure~\ref{fig:arch}}
	\newcommand{\reffigGDarch}{Figure~\ref{fig:GDarch}}
	\newcommand{\refeqmodulation}{Equation~\ref{eq:modulation}}
	\newcommand{\refeqdemodulation}{Equation~\ref{eq:demodulation}}
	\newcommand{\refeqmodem}{Equations \ref{eq:modulation} and \ref{eq:demodulation}}
	\newcommand{\refseclazyreg}{Section~\ref{sec:lazyreg}}
	\newcommand{\refsecpathreg}{Section~\ref{sec:pathreg}}
	\newcommand{\refsecaltarch}{Section~\ref{sec:altarch}}
	\newcommand{\reftabmain}{Table~\ref{tab:main}}
	\newcommand{\reftabmains}{Tables \ref{tab:main} and \ref{tab:lsun}}
\else
	\newcommand{\refsecin}{Section~2 of the paper}
	\newcommand{\reffigarch}{Figure~1 of the paper}
	\newcommand{\reffigGDarch}{Figure~7 of the paper}
	\newcommand{\refeqmodulation}{Equation~1 of the paper}
	\newcommand{\refeqdemodulation}{Equation~3 of the paper}
	\newcommand{\refeqmodem}{Equations 1 and 3 of the paper}
	\newcommand{\refseclazyreg}{Section~3.1 of the paper}
	\newcommand{\refsecpathreg}{Section~3.2 of the paper}
	\newcommand{\refsecaltarch}{Section~4.1 of the paper}
	\newcommand{\reftabmain}{Table~1 of the paper}
	\newcommand{\reftabmains}{Tables~1 and~3 of the paper}
	\newcommand{\reftablsun}{Table~3 of the paper}
\fi

\section{Image quality} %
\label{app:imagequality}

\figffhqcherrypick %
\figuncurated %
\figcatscomparison %
\figcarscomparison %
\figartifactsapp
\figfrequencies

We include several large images that illustrate various aspects related to image quality.
Figure~\ref{fig:ffhqcherrypick} shows hand-picked examples illustrating the quality and diversity achievable using our method in \textsc{FFHQ}, while
Figure~\ref{fig:uncurated} shows uncurated results for all datasets mentioned in the paper.

Figures~\ref{fig:catscomparison} and~\ref{fig:carscomparison} demonstrate cases where FID and P\&R give non-intuitive results, but PPL seems to be more in line with human judgement.

We also include images relating to StyleGAN artifacts.
Figure~\ref{fig:artifactsapp} shows a rare case where the blob artifact fails to appear in StyleGAN activations, leading to a seriously broken image.
Figure~\ref{fig:frequencies} visualizes the activations inside \reftabmain{} configurations~\arch{a} and~\arch{f}. It is evident that progressive growing leads to higher-frequency content in the intermediate layers, compromising shift invariance of the network. We hypothesize that this causes the observed uneven location preference for details when progressive growing is used.

\section{Implementation details} %
\label{app:implementation}

We implemented our techniques on top of the official TensorFlow implementation of StyleGAN\footnote{\texttt{https://github.com/NVlabs/stylegan}} corresponding to configuration \arch{a} in \reftabmain{}.
We kept most of the details unchanged, including
	the dimensionality of $\ZZ$ and $\WW$ (512),
	mapping network architecture (8 fully connected layers, 100$\times$ lower learning rate),
	equalized learning rate for all trainable parameters \cite{Karras2017},
	leaky ReLU activation with $\alpha=0.2$,
	bilinear filtering \cite{zhang2019} in all up/downsampling layers \cite{Karras2018},
	minibatch standard deviation layer at the end of the discriminator \cite{Karras2017},
	exponential moving average of generator weights \cite{Karras2017},
	style mixing regularization \cite{Karras2018},
	non-saturating logistic loss \cite{Goodfellow2014} with $R_1$ regularization \cite{Mescheder2018},
	Adam optimizer \cite{adam} with the same hyperparameters ($\beta_1=0, \beta_2=0.99, \epsilon=10^{-8}, \mathrm{minibatch}=32$),
	and training datasets \cite{Karras2018,LSUN}.
We performed all training runs on NVIDIA DGX-1 with 8 Tesla V100 GPUs using TensorFlow 1.14.0 and cuDNN 7.4.2.

\paragraph{Generator redesign}
In configurations \arch{b}--\arch{f} we replace the original StyleGAN generator with our revised architecture.
In addition to the changes highlighted in \refsecin{}, we initialize components of the constant input $c_1$ using $\mathcal{N}(0,1)$ and simplify the noise broadcast operations to use a single shared scaling factor for all feature maps.
\FINAL{Similar to Karras~et~al.~\cite{Karras2018}, we initialize all weights using $\mathcal{N}(0,1)$ and all biases and noise scaling factors to zero, except for the biases of the affine transformation layers, which we initialize to one.}
We employ weight modulation and demodulation in all convolution layers, except for the output layers (tRGB in \reffigGDarch{}) where we leave out the demodulation.
With $1024^2$ output resolution, the generator contains a total of 18 affine transformation layers where the first one corresponds to $4^2$ resolution, the next two correspond to $8^2$, and so forth.

\paragraph{Weight demodulation}
Considering the practical implementation of \refeqmodem{}, it is important to note that the resulting set of weights will be different for each sample in a minibatch, which rules out direct implementation using standard convolution primitives.
Instead, we choose to employ \emph{grouped convolutions} \cite{krizhevsky2012} that were originally proposed as a way to reduce computational costs by dividing the input feature maps into multiple independent groups, each with their own dedicated set of weights.
We implement \refeqmodem{} by temporarily reshaping the weights and activations so that each convolution sees one sample with $N$ groups\,---\,instead of $N$ samples with one group.
This approach is highly efficient because the reshaping operations do not actually modify the contents of the weight and activation tensors.

\paragraph{Lazy regularization}
In configurations \arch{c}--\arch{f} we employ lazy regularization (\refseclazyreg{}) by evaluating the regularization terms ($R_1$ and path length) in a separate regularization pass that we execute once every $k$ training iterations.
We share the internal state of the Adam optimizer between the main loss and the regularization terms, so that the optimizer first sees gradients from the main loss for $k$ iterations, followed by gradients from the regularization terms for one iteration.
To compensate for the fact that we now perform $k+1$ training iterations instead of $k$, we adjust the optimizer hyperparameters $\lambda' = c\cdot\lambda$, $\beta_1' = (\beta_1)^c$, and $\beta_2' = (\beta_2)^c$, where $c = k/(k+1)$.
We also multiply the regularization term by $k$ to balance the overall magnitude of its gradients.
We use $k=16$ for the discriminator and $k=8$ for the generator.

\paragraph{Path length regularization}
Configurations \arch{d}--\arch{f} include our new path length regularizer (\refsecpathreg{}).
We initialize the target scale $a$ to zero and track it on a per-GPU basis as the exponential moving average of $\left\lVert \mathbf{J}_{\ww}^T \yy\right\rVert_2$ using decay coefficient $\beta_\mathrm{pl} = 0.99$.
We weight our regularization term by
\begin{equation}
\gamma_\textrm{pl} = \frac{\ln 2}{r^2 (\ln r - \ln 2)}
\textrm{,}
\end{equation}
where $r$ specifies the output resolution (e.g. $r=1024$).
We have found these parameter choices to work reliably across all configurations and datasets.
To ensure that our regularizer interacts correctly with style mixing regularization, we compute it as an average of all individual layers of the synthesis network. %
Appendix~\ref{app:regularization} provides detailed analysis of the effects of our regularizer on the mapping between $\WW$ and image space.

\paragraph{Progressive growing}
In configurations \arch{a}--\arch{d} we use progressive growing with the same parameters as Karras~et~al.~\cite{Karras2018} (start at $8^2$ resolution and learning rate $\lambda = 10^{-3}$, train for 600k images per resolution, fade in next resolution for 600k images, increase learning rate gradually by $3\times$).
In configurations \arch{e}--\arch{f} we disable progressive growing and set the learning rate to a fixed value $\lambda = 2 \cdot 10^{-3}$, which we found to provide the best results.
In addition, we use output skips in the generator and residual connections in the discriminator as detailed in \refsecaltarch{}.

\paragraph{Dataset-specific tuning}
Similar to Karras~et~al.~\cite{Karras2018}, we augment the \textsc{FFHQ} dataset with horizontal flips to effectively increase the number of training images from 70k to 140k, and we do not perform any augmentation for the \textsc{LSUN} datasets.
We have found that the optimal choices for the training length and $R_1$ regularization weight $\gamma$ tend to vary considerably between datasets and configurations.
We use $\gamma = 10$ for all training runs except for configuration \arch{e} in \reftabmain{}, as well as \textsc{LSUN Church} and \textsc{LSUN Horse} in \reftablsun{}, where we use $\gamma = 100$.
It is possible that further tuning of $\gamma$ could provide additional benefits.

\paragraph{Performance optimizations}
We profiled our training runs extensively and found that\,---\,in our case\,---\,the default primitives for image filtering, up/downsampling, bias addition, and leaky ReLU had surprisingly high overheads in terms of training time and GPU memory footprint.
This motivated us to optimize these operations using hand-written CUDA kernels.
We implemented filtered up/downsampling as a single fused operation, and bias and activation as another one.
In configuration \arch{e} at $1024^2$ resolution, our optimizations improved the overall training time by about 30\% and memory footprint by about 20\%. %

\section{Effects of path length regularization}  %
\label{app:regularization}

The path length regularizer described in \refsecpathreg{} is of the form:

\begin{equation}
	\mathcal{L}_{\mathrm{pl}} = \expectation_\ww \expectation_\yy \left(\left\lVert \mathbf{J}_{\ww}^T \yy\right\rVert_2 - a\right)^2,
	\label{eq:appxppl}
\end{equation}
where $\yy \in \real^{M}$ is a unit normal distributed random variable in the space of generated images (of dimension $M = 3wh$, namely the RGB image dimensions), $\JJ_\ww \in \real^{M \times L}$ is the Jacobian matrix of the generator function $g: \real^{L} \mapsto \real^{M}$ at a latent space point $\ww \in \real^L$, and $a \in \mathbb{R}$ is a global value that expresses the desired scale of the gradients.

\subsection{Effect on pointwise Jacobians}
The value of this prior is minimized when the inner expectation over $\yy$ is minimized at every latent space point $\ww$ separately. In this subsection, we show that the inner expectation is (approximately) minimized when the Jacobian matrix $\JJ_\ww$ is orthogonal, up to a global scaling factor. The general strategy is to use the well-known fact that, in high dimensions $L$, the density of a unit normal distribution is concentrated on a spherical shell of radius $\sqrt{L}$. The inner expectation is then minimized when the matrix $\JJT_\ww$ scales the function under expectation to have its minima at this radius. This is achieved by any orthogonal matrix (with suitable global scale that is the same at every $\ww$).

We begin by considering the inner expectation
\begin{equation*}
	\mathcal{L}_{\ww} := \expectation_\yy \left(\left\lVert \JJT_\ww \yy\right\rVert_2 - a\right)^2.
\end{equation*}
We first note that the radial symmetry of the distribution of $\yy$, as well as of the $l_2$ norm, allows us to focus on diagonal matrices only. This is seen using the Singular Value Decomposition $\JJT_\ww = \mathbf{U}\tilde\SSigma\mathbf{V}^T$, where $\mathbf{U} \in \real^{L \times L}$ and $\mathbf{V} \in \real^{M \times M}$ are orthogonal matrices, and $\tilde \SSigma = [\SSigma ~ \mathbf{0}]$ is a horizontal concatenation of a diagonal matrix $\SSigma \in \real^{L \times L}$ and a zero matrix $\mathbf{0} \in \real^{L \times (M-L)}$ \cite{Golub2013}. Because rotating a unit normal random variable by an orthogonal matrix leaves the distribution unchanged, and rotating a vector leaves its norm unchanged, the expression simplifies to
\begin{eqnarray*}
	\mathcal{L}_{\ww} &=& \expectation_\yy \left(\left\lVert \mathbf{U}\tilde\SSigma\mathbf{V}^T \yy\right\rVert_2 - a\right)^2 
	\\
	&=& \expectation_\yy \left(\left\lVert \tilde\SSigma \yy\right\rVert_2 - a\right)^2.
\end{eqnarray*}
Furthermore, the zero matrix in $\tilde \SSigma$ %
drops the dimensions of $\yy$ beyond $L$, effectively marginalizing its distribution over those dimensions. The marginalized distribution is again a unit normal distribution over the remaining $L$ dimensions. We are then left to consider the minimization of the expression
\begin{equation*}
	\mathcal{L}_{\ww} = \expectation_{\tyy} \left(\left\lVert \SSigma \tyy\right\rVert_2 - a\right)^2,
\end{equation*}
over diagonal square matrices $\SSigma \in \real^{L \times L}$, where $\tyy$ is unit normal distributed in dimension $L$. To summarize, all matrices $\JJT_\ww$ that share the same singular values with $\SSigma$ produce the same value for the original loss.

Next, we show that this expression is minimized when the diagonal matrix $\SSigma$ has a specific identical value at every diagonal entry, i.e., it is a constant multiple of an identity matrix. We first write the expectation as an integral over the probability density of $\tyy$:
\begin{eqnarray*}
	\mathcal{L}_{\ww} &=& \int \left(\left\lVert \SSigma \tyy\right\rVert_2 - a\right)^2 p_{\tyy}(\tyy) ~ \mathrm{d}\tyy \\
	&=& (2 \pi)^{-\frac{L}{2}} \int \left(\left\lVert \SSigma \tyy\right\rVert_2 - a\right)^2 \mathrm{exp}\left(-\frac{\tyy^T \tyy}{2}\right) ~ \mathrm{d}\tyy 
\end{eqnarray*}
Observing the radially symmetric form of the density, we change into a polar coordinates $\tyy = r \phi$, where $r \in \real_+$ is the distance from origin, and $\phi \in \mathbb{S}^{L-1}$ is a unit vector, i.e., a point on the $L-1$-dimensional unit sphere. This change of variables introduces a Jacobian factor $r^{L-1}$:
\begin{multline*}
	\tilde{\mathcal{L}}_{\ww} =
	(2 \pi)^{-\frac{L}{2}} \int_{\mathbb{S}}\int_0^\infty \left(r\left\lVert \SSigma \phi \right\rVert_2 - a\right)^2 r^{L-1} \\
	\mathrm{exp}\left(-\frac{r^2}{2}\right) ~ \mathrm{d}r ~ \mathrm{d}\phi
\end{multline*}

The probability density $(2\pi)^{-L/2} r^{L-1} \mathrm{exp}\left(-\frac{r^2}{2}\right)$ is then an $L$-dimensional unit normal density expressed in polar coordinates, dependent only on the radius and not on the angle. A standard argument by Taylor approximation shows that when $L$ is high, for any $\phi$ the density is well approximated by density $(2\pi e / L)^{-L/2} \mathrm{exp}\left(-\frac{1}{2} (r-\mu)^2 / \sigma^2\right)$, which is a (unnormalized) one-dimensional normal density in $r$, centered at $\mu=\sqrt{L}$ of standard deviation $\sigma = 1/\sqrt{2}$ \cite{Bishop2006}. In other words, the density of the $L$-dimensional unit normal distribution is concentrated on a shell of radius $\sqrt{L}$. Substituting this density into the integral, the loss becomes approximately
\begin{multline}
	\mathcal{L}_{\ww} \approx (2\pi e / L)^{-L/2} \int_{\mathbb{S}}\int_0^\infty  \left(r\left\lVert \SSigma \phi \right\rVert_2 - a\right)^2 \\
	 \mathrm{exp}\left(-\frac{\left(r-\sqrt{L}\right)^2}{2 \sigma^2}\right) ~ \mathrm{d}r ~ \mathrm{d}\phi,
	\label{eq:gaussshell}
\end{multline}
where the approximation becomes exact in the limit of infinite dimension $L$.

To minimize this loss, we set $\SSigma$ such that the function $\left(r\left\lVert \SSigma \phi \right\rVert_2 - a\right)^2$ obtains minimal values on the spherical shell of radius $\sqrt{L}$. This is achieved by $\SSigma = \frac{a}{\sqrt{L}} \mathbf{I}$, whereby the function becomes constant in $\phi$ and the expression reduces to
\begin{multline*}
	\mathcal{L}_{\ww} \approx (2\pi e/L)^{-L/2} \mathcal{A}(\mathbb{S}) a^2 L^{-1} \int_0^\infty \left(r - \sqrt{L}\right)^2 \\ 
	\mathrm{exp}\left(-\frac{\left(r-\sqrt{L}\right)^2}{2 \sigma^2}\right) ~ \mathrm{d}r,
\end{multline*}
where $\mathcal{A}(\mathbb{S})$ is the surface area of the unit sphere (and like the other constant factors, irrelevant for minimization). Note that the zero of the parabola $(r-\sqrt{L})^2$ coincides with the maximum of the probability density, and therefore this choice of $\Sigma$ minimizes the inner integral in Eq.~\ref{eq:gaussshell} separately for every $\phi$.

In summary, we have shown that\,---\,assuming a high dimensionality $L$ of the latent space\,---\,the value of the path length prior (Eq.~\ref{eq:appxppl}) is minimized when all singular values of the Jacobian matrix of the generator are equal to a global constant, at every latent space point $\ww$, i.e., they are orthogonal up to a globally constant scale.

While in theory $a$ merely scales the values of the mapping without changing its properties and could be set to a fixed value (e.g., $1$), in practice it does affect the dynamics of the training. If the imposed scale does not match the scale induced by the random initialization of the network, the training spends its critical early steps in pushing the weights towards the required overall magnitudes, rather than enforcing the actual objective of interest. This may degrade the internal state of the network weights and lead to sub-optimal performance in later training. Empirically we find that setting a fixed scale reduces the consistency of the training results across training runs and datasets. Instead, we set $a$ dynamically based on a running average of the existing scale of the Jacobians, namely $a \approx \expectation_{\ww,\yy} \left(\left\lVert \JJT_w \yy\right\rVert_2\right)$. With this choice the prior targets the scale of the local Jacobians towards whatever global average already exists, rather than forcing a specific global average. This also eliminates the need to measure the appropriate scale of the Jacobians explicitly, as is done by Odena et al.~\cite{Odena2018} who consider a related conditioning prior.

\figeigenvalues

Figure \ref{fig:eigenvalues} shows empirically measured magnitudes of singular values of the Jacobian matrix for networks trained with and without path length regularization. While orthogonality is not reached, the eigenvalues of the regularized network are closer to one another, implying better conditioning, with the strength of the effect correlated with the PPL metric (\reftabmain{}).

\subsection{Effect on global properties of generator mapping}
In the previous subsection, we found that the prior encourages the Jacobians of the generator mapping to be everywhere orthogonal. While Figure~\ref{fig:eigenvalues} shows that the mapping does not satisfy this constraint exactly in practice, it is instructive to consider what global properties the constraint implies for mappings that do.  Without loss of generality, we assume unit global scale for the matrices to simplify the presentation.

The key property is that that a mapping $g: \real^L \mapsto \real^M$ with everywhere orthogonal Jacobians preserves the lengths of curves. To see this, let $u: [t_0, t_1] \mapsto \real^L$ parametrize a curve in the latent space. Mapping the curve through the generator $g$, we obtain a curve $\tilde u = g \circ u$ in the space of images. Its arc length is
\begin{equation}
	L = \int_{t_0}^{t_1} \lvert \tilde u'(t) \rvert ~\mathrm{d}t,
\end{equation}
where prime denotes derivative with respect to $t$. By chain rule, this equals
\begin{equation}
	L = \int_{t_0}^{t_1} \lvert J_g(u(t)) u'(t) \rvert ~\mathrm{d}t,
\end{equation}
where $J_g \in \real^{L \times M}$ is the Jacobian matrix of $g$ evaluated at $u(t)$. By our assumption, the Jacobian is orthogonal, and consequently it leaves the 2-norm of the vector $u'(t)$ unaffected:
\begin{equation}
	L = \int_{t_0}^{t_1} \lvert u'(t) \rvert ~\mathrm{d}t.
\end{equation}
This is the length of the curve $u$ in the latent space, prior to mapping with $g$. Hence, the lengths of $u$ and $\tilde u$ are equal, and so $g$ preserves the length of any curve.

In the language of differential geometry, $g$ isometrically embeds the Euclidean latent space $\real^L$ into a submanifold $\mathcal{M}$ in $\real^M$\,---\,e.g., the manifold of images representing faces, embedded within the space of all possible RGB images. A consequence of isometry is that straight line segments in the latent space are mapped to geodesics, or shortest paths, on the image manifold: a straight line $v$ that connects two latent space points cannot be made any shorter, so neither can there be a shorter on-manifold image-space path between the corresponding images than $g \circ v$. For example, a geodesic on the manifold of face images is a continuous morph between two faces that incurs the minimum total amount of change (as measured by $l_2$ difference in RGB space) when one sums up the image difference in each step of the morph.

Isometry is not achieved in practice, as demonstrated in empirical experiments in the previous subsection. The full loss function of the training is a combination of potentially conflicting criteria, and it is not clear if a genuinely isometric mapping would be capable of expressing the image manifold of interest. Nevertheless, a pressure to make the mapping as isometric as possible has desirable consequences. In particular, it discourages unnecessary ``detours'': in a non-constrained generator mapping, a latent space interpolation between two similar images may pass through any number of distant images in RGB space. With regularization, the mapping is encouraged to place distant images in different regions of the latent space, so as to obtain short image paths between any two endpoints.

\section{Projection method details} %
\label{app:projectiondetails}

\fignoiseregcars{} %

\newcommand{\imgsym}{\boldsymbol{x}}
\newcommand{\noisesym}{\boldsymbol{n}}
Given a target image $\imgsym$, we seek to find the corresponding $\ww \in \WW$ and per-layer noise maps denoted $\noisesym_i \in \real^{r_i \times r_i}$ where $i$ is the layer index and $r_i$ denotes the resolution of the $i$th noise map.
The baseline StyleGAN generator in 1024$\times$1024 resolution has 18 noise inputs, i.e., two for each resolution from 4$\times$4 to 1024$\times$1024 pixels.
Our improved architecture has one fewer noise input because we do not add noise to the learned 4$\times$4 constant (\reffigarch{}).

Before optimization, we compute $\MU_\ww = \expectation_\zz\,f(\zz)$ by running 10\,000 random latent codes $\zz$ through the mapping network $f$.
We also approximate the scale of $\WW$ by computing $\sigma^2_\ww = \expectation_\zz\,\lVert f(\zz)-\MU_\ww\rVert^2_2$, i.e., the average square Euclidean distance to the center.

At the beginning of optimization, we initialize $\ww = \MU_\ww$ and $\noisesym_i = \mathcal{N}(\boldsymbol{0}, \II)$ for all $i$. 
The trainable parameters are the components of $\ww$ as well as all components in all noise maps $\noisesym_i$.
The optimization is run for 1000 iterations using Adam optimizer~\cite{adam} with default parameters.
Maximum learning rate is $\lambda_\mathit{max}=0.1$, and it is ramped up from zero linearly during the first 50 iterations and ramped down to zero using a cosine schedule during the last 250 iterations.
In the first three quarters of the optimization we add Gaussian noise to $\ww$ when evaluating the loss function 
as $\tildeww = \ww + \mathcal{N}(0, 0.05\,\sigma_\ww t^2)$, where $t$ goes from one to zero during the first 750 iterations.
This adds stochasticity to the optimization and stabilizes finding of the global optimum.

Given that we are explicitly optimizing the noise maps, we must be careful to avoid the optimization from sneaking actual signal into them.
Thus we include several noise map regularization terms in our loss function, in addition to an image quality term. 
The image quality term is the LPIPS~\cite{Zhang2018metric} distance between target image $\imgsym$ and the synthesized image:
$L_\mathit{image}=D_\mathrm{LPIPS}[\imgsym, g(\tildeww, \noisesym_0, \noisesym_1, \ldots)]$.
For increased performance and stability, we downsample both images to 256$\times$256 resolution before computing the LPIPS distance.
Regularization of the noise maps is performed on multiple resolution scales.
For this purpose, we form for each noise map greater than 8$\times$8 in size a pyramid down to 8$\times$8 resolution by averaging 2$\times$2 pixel neighborhoods and multiplying by 2 at each step to retain the expected unit variance. 
These downsampled noise maps are used for regularization only and have no part in synthesis.

Let us denote the original noise maps by $\noisesym_{i,0} = \noisesym_{i}$ and the downsampled versions by $\noisesym_{i,j>0}$.
Similarly, let $r_{i,j}$ be the resolution of an original ($j=0$) or downsampled ($j>0$) noise map so that $r_{i,j+1} = r_{i,j}/2$.
The regularization term for noise map $\noisesym_{i,j}$ is then
\begin{eqnarray*}
L_{i,j} &=& \left( \frac{1}{r_{i,j}^2} \cdot \sum_{x,y}\noisesym_{i,j}(x,y)\cdot\noisesym_{i,j}(x-1,y) \right)^2 \\
        &+& \left( \frac{1}{r_{i,j}^2} \cdot \sum_{x,y}\noisesym_{i,j}(x,y)\cdot\noisesym_{i,j}(x,y-1) \right)^2
\textrm{,}
\end{eqnarray*}
where the noise map is considered to wrap at the edges.
The regularization term is thus sum of squares of the resolution-normalized autocorrelation coefficients at one pixel shifts horizontally and vertically, which should be zero for a normally distributed signal.
The overall loss term is then $L_\mathit{total}=L_\mathit{image}+\alpha\sum_{i,j}L_{i,j}$. In all our tests, we have used noise regularization weight $\alpha=10^5$.
In addition, we renormalize all noise maps to zero mean and unit variance after each optimization step. %
Figure~\ref{fig:noiseregcars} illustrates the effect of noise regularization on the resulting noise maps.

\section{\FINAL{Results with spectral normalization}}
\label{app:spectralnorm}

\tabspectralnorm

\FINAL{%
Since spectral normalization (SN) is widely used in GANs \cite{Miyato2018B}, we investigated its effect on StyleGAN2.
Table~\ref{tab:spectralnorm} gives the results for a variety of configurations where spectral normalization is enabled in addition to our techniques (weight demodulation, path length regularization) or instead of them.

Interestingly, adding spectral normalization to our generator is almost a no-op. On an implementation level, SN scales the weight tensor of each layer with a scalar value $1/\sigma(w)$. The effect of such scaling, however, is overridden by \refeqdemodulation{} for the main convolutional layers as well as the affine transformation layers. Thus, the only thing that SN adds on top of weight demodulation is through its effect on the \texttt{tRGB} layers.

When we enable spectral normalization in the discriminator, FID is slightly compromised. Enabling it in the generator as well leads to significantly worse results, even though its effect is isolated to the \texttt{tRGB} layers. Leaving SN enabled, but disabling a subset of our contributions does not improve the situation. 
Thus we conclude that StyleGAN2 gives better results without spectral normalization.
}

\section{\FINAL{Energy consumption}}
\label{app:power}

\tabpower

\FINAL{
Computation is a core resource in any machine learning project: its availability and cost, as well as the associated energy consumption, are key factors in both choosing research directions and practical adoption. We provide a detailed breakdown for our entire project in Table~\ref{tab:power} in terms of both GPU time and electricity consumption.

We report expended computational effort as single-GPU years (Volta class GPU). We used a varying number of NVIDIA DGX-1s for different stages of the project, and converted each run to single-GPU equivalents by simply scaling by the number of GPUs used.

The entire project consumed approximately 131.61 megawatt hours (MWh) of electricity. We followed the Green500 power measurements guidelines \cite{Ge2020} as follows. 
For each job, we logged the exact duration, number of GPUs used, and which of our two separate compute clusters the job was executed on. We then measured the actual power draw of an 8-GPU DGX-1 when it was training \textsc{FFHQ} config~\arch{f}. A separate estimate was obtained for the two clusters because they use different DGX-1 SKUs. The vast majority of our training runs used 8 GPUs, and for the rest we approximated the power draw by scaling linearly with $n/8$, where $n$ is the number of GPUs. 

Approximately half of the total energy was spent on early exploration and forming ideas. Then subsequently a quarter was spent on refining those ideas in more targeted experiments, and finally a quarter on producing this paper and preparing the public release of code, trained models, and large sets of images. Training a single \textsc{FFHQ} network (config~\arch{f}) took approximately 0.68~MWh (0.5\% of the total project expenditure). This is the cost that one would pay when training the network from scratch, possibly using a different dataset. In short, vast majority of the electricity used went into shaping the ideas, testing hypotheses, and hyperparameter tuning. We did not use automated tools for finding hyperparameters or optimizing network architectures.
}

\fi

\end{document}